\documentclass[journal,comsoc]{IEEEtran}

\usepackage{amssymb}
\usepackage{amsthm,amssymb}
\usepackage{eucal}

\usepackage{cite}
\usepackage{amsmath,amssymb,amsfonts}
\usepackage{amsmath}
\usepackage{algorithmic}
\usepackage{graphicx}
\usepackage{textcomp}
\usepackage{xcolor}
\usepackage{lettrine}
 \usepackage{multirow}
 \usepackage{amsmath}
 \usepackage{graphicx}
 \usepackage{textgreek}
\usepackage{amssymb}
\usepackage{caption}
\usepackage{subcaption}
\usepackage{authblk}

\newcommand{\RN}[1]{%
  \textup{\uppercase\expandafter{\romannumeral#1}}%
}
\usepackage[ruled,longend]{algorithm2e}
\def\BibTeX{{\rm B\kern-.05em{\sc i\kern-.025em b}\kern-.08em
    T\kern-.1667em\lower.7ex\hbox{E}\kern-.125emX}}
\def\BibTeX{{\rm B\kern-.05em{\sc i\kern-.025em b}\kern-.08em
    T\kern-.1667em\lower.7ex\hbox{E}\kern-.125emX}}
\newtheorem{proposition}{\textbf{Proposition}}

\newtheorem{lemma}{\textbf{Lemma}}

\begin{document}

\title{Reinforcement Learning for Deceiving  Reactive  Jammers in Wireless Networks}
\IEEEoverridecommandlockouts

  \author{\IEEEauthorblockN{\normalsize Ali Pourranjbar, Georges Kaddoum,\textit{ IEEE senior
Member}, Aidin Ferdowsi, \textit{ IEEE Student Member,}
and Walid Saad, \textit{ IEEE Fellow} 
}

\thanks{
A. Pourranjbar and G. Kaddoum are with the LaCIME Lab, Department of Electrical Engineering, École de technologie supérieure, Montreal,
QC H3C 0J9, Canada (e-mail: ali.pourranjbar.1@ens.etsmtl.ca; Georges.Kaddoum@etsmtl.ca).

A. Ferdowsi and W. Saad are with Wireless@VT, Bradley Department of Electrical and Computer Engineering, Virginia Tech, Blacksburg 24061,VA USA (e-mail: aidin@vt.edu; walids@vt.edu).}
 }
\maketitle

 \vspace{-0.5cm}
\begin{abstract}
Conventional anti-jamming methods  mostly rely on frequency hopping to hide or  escape   from jammers. These approaches are not efficient in terms of bandwidth usage and can also
result in a high probability of jamming. Different from existing works, in this paper, a novel anti-jamming strategy is proposed based on the idea of deceiving the jammer into attacking a victim channel while maintaining  the communications of  legitimate users  in safe channels.  Since the jammer’s  channel information is not known to the users,  an optimal channel selection scheme and a sub-optimal power allocation algorithm are proposed using reinforcement learning (RL). The performance of the proposed anti-jamming technique is evaluated by deriving the statistical lower bound of  the  total received power (TRP). Analytical results show that, for a given access point, over $50\%$ of the highest achievable TRP, i.e.  in the absence of jammers, is achieved  for the case of a single user and three frequency channels. Moreover, this value increases with  the number of users and available channels. The obtained results are compared with  two existing  RL based anti-jamming techniques, and a random channel allocation strategy without any jamming attacks.
Simulation results show that the proposed anti-jamming method outperforms the compared RL based anti-jamming methods and the random search method, and yields near optimal achievable TRP.  

\end{abstract}

\begin{IEEEkeywords}
 Reactive jammer, frequency hopping, reinforcement learning, deception.
\end{IEEEkeywords}
 \vspace{-0.3cm}
\section{Introduction}
 \lettrine[lines=2]{W}{}ireless  communication networks are known to be vulnerable to malicious attacks such as jamming\cite{Paper1}. Jammers mostly impact  the physical layer  by transmitting disruptive signals over shared wireless communication channels. Under   jamming attacks, wireless  network  components  are supposed to  consume  more power or retransmit  the lost data to compensate the jamming effects. The former strategy is energy inefficient while the latter can significantly decrease the data rate. Thus, to maintain an adequate quality-of-service (QoS), anti-jamming policies are needed. Jammers are typically classified based on their jamming policies from elementary to  advanced jammers \cite{Paper2}. Elementary  jammers adopt a predefined technique, such as   constant, random, and sweeping jammers. Advanced jammers adapt jamming techniques based on the opponent's actions. For example, reactive jammers select their power and channel  according to their opponents' channels and power levels. 
 
For elementary  jammers, once their polices are detected, jamming mitigation can be performed by frequency band or power adaptation. However, behavior of  advanced jammers should be monitored in order to mitigate the jamming effect.  Anti-jamming methods should be designed such that the communication resource usage can be optimized  while mitigating the jamming effects.  
\vspace{-0.2cm}
\subsection{Related Works}
 
 Numerous anti-jamming methods have been proposed in the literature, ranging from frequency hopping  \cite{Paper3,Paper4,Paper5,Paper6,Paper7,Paper8} methods that employ  techniques such as honeypots to obtain the jammer policy or to harvest the jamming energy \cite{Paper9,Paper10,Paper11,Paper12,Paper13,Paper14}.  Frequency hopping methods continuously switch the carrier frequency between different bands and can be performed using strategies such as chaotic frequency hopping \cite{Paper7 } or learning-based methods \cite{Paper8}. The authors in \cite{Paper9} propose an anti-jamming technique that assigns a user among all users as a honeypot to obtain the jammer policy for a wiser jamming mitigation. The work in\cite{Paper10} proposes an anti-jamming method
based on dispersing the data in time frames, and models  the impacts of the spectrum changes on the mobile cognitive users' performance in hostile environments.  In \cite{Paper11},  the authors introduce a multi-domain anti-jamming method that uses both of the frequency and power domains to overcome smart jammer attacks. The authors in \cite{Paper12} employ  an  unmanned aerial vehicle (UAV) to hold a communication link between a user and a backup base station when the  communication link with the main base station is disrupted. The work in \cite{Paper13} proposes a collaborative anti-jamming algorithm (CMAA) in which users   collaborate with each other in terms of frequency channel selection in order to mitigate the jammer's effects.
 \textcolor{black}{In \cite{Paper14}, the authors  propose an spectrum sensing based anti-jamming method where legitimate users mitigate the  jamming effects by enhancing their awareness  about the jammed channels. }

A number of prior works   developed anti-jamming techniques based on game theory\cite{Paper15,Paper16,Paper17,Paper18,Paper19,Paper20}. The authors in \cite{Paper15} propose  a noncooperative game to select the optimum relay station   in the presence of an adversary. 
 The authors in \cite{Paper16} seek to mitigate the jammer effect in an OFDM-based Internet of Things system by dispersing an access point (AP) power among sub-carriers. In \cite{Paper17} and \cite{Paper1}, the authors study the impact of  the observation  error of the legitimate users and jammers on the network performance, respectively. In \cite{Paper20}, the authors propose a dynamic game to deceive a jammer in a cooperative drone scenario.

In \cite{Paper21,Paper22,Paper23,Paper24,Paper25,Paper26,Paper27}, machine learning-based anti-jamming techniques are proposed. In \cite{Paper21}, the authors employ a deep Q-learning learning (DQL) based anti-jamming method  to  mitigate  the  effects  of  a  powerful Markov jammer. The work in \cite{Paper22} proposes a deep reinforcement learning (RL) based anti-jamming technique against a smart jammer in a non-orthogonal multiple access system. In \cite{Paper23}, the authors employ  deep RL (DRL) to secure the communication between a transmitter and a receiver against multi-jammers. The work in\cite{Paper24} proposes a  modified Q-learning technique, where  all the Q-values of the Q-table are updated at each iteration, to mitigate the effects of a sweeping jammer. A DRL based method to  obtain the optimal task offloading policy under jamming attacks in the context of multi-radio access   is proposed  in \cite{Paper25}. 
Authors in \cite{Paper26} propose the idea of  harvesting the transmitted power by  jammers for data transmission.
The work in \cite{Paper27} introduces a system consisting of two groups of nodes, namely legitimate users and jammers, that compete to dominate the shared spectrum. In this regard, multi-agent Q-learning is employed to discover the optimal actions of the nodes.
 The works in  \cite{Paper28,Paper29,Paper30,Paper31} develop anti-jamming methods that employ new approaches to deceive the jammer using a honeypot or fake transmission.
The work in \cite{Paper28} proposes an  anti-jamming algorithm  in which ``decoy'' users are used to trap the jammer. Similar to  \cite{Paper28}, the authors in  \cite{Paper29} propose   to foil the jammer by dedicating a secondary user that transmits  fake signals to attract a portion of the jamming power.  The authors in \cite{Paper30} propose an anti-jamming method where a transmitter forms a decoy beam in another frequency channel than the main communication channel  to distract the jammer from the main communication beam. Inspired by  \cite{Paper26}, the work in\cite{Paper31}    employs a radio  frequency (RF) tag that uses  the harvested energy from the jamming signal to back scatter the transmitter information to a multi-array receiver while the transmitter keeps  the main transmission to deceive the jammer.

Despite their position in the spotlight when it comes to the mitigation of jamming attacks, frequency hopping based anti-jamming methods are not efficient  in terms of bandwidth usage and can also  result in a high probability of jamming\cite{Paper3,Paper4,Paper5,Paper6,Paper7,Paper8,Paper21} and \cite{Paper27}.  Moreover, the channel qualities are most often neglected  in  frequency hopping based methods. Some works such as \cite{Paper11}, \cite{Paper12}, \cite{Paper15}, \cite{Paper16}, \cite{Paper20}, \cite{Paper26}, and \cite{Paper27}  address this problem; however, in \cite{Paper11} and \cite{Paper27}   full knowledge of the environment is assumed to be available and  the proposed anti-jamming methods by \cite{Paper12}, \cite{Paper15}, \cite{Paper16}, \cite{Paper20}, and  \cite{Paper26} are restricted to a certain considered system model. For instance,  jamming effects are mitigated in UAV-based systems in \cite{Paper15}  and in scenario where the users are able to harvest energy in \cite{Paper26}. In addition, the necessity of channel switching in frequency hopping methods causes  communication delay  and energy consumption\cite{Paper10}.
Considering a jammer with  simple jamming policy  as the opponent is another drawback of previous
works such as \cite{Paper23} and \cite{Paper24}, which makes their proposed anti-jamming methods impractical in
realistic scenarios where jammers are more developed. For instance, the considered intelligent jammer in \cite{Paper23} selects three channels and keeps jamming those channels for  a  specific  amount  of  time while the reactive jammer in \cite{Paper24}   jams the sensed channel after two time slots. These types of jammers' policies can be easily detected by monitoring their behavior during a short period of time.

Although the deception techniques proposed in \cite{Paper28,Paper29,Paper30,Paper31} can reduce  the channel switching rate using a decoy or fake transmission to trap the jammer,  they   have  a number of drawbacks. These works  assume   full knowledge of the environment is available, which is not a practical assumption   due the unpredictable nature of  jammers. As a result, the problem of finding the optimal channel allocation and user selection as a decoy are not considered. Moreover, since the works in \cite{Paper28} and \cite{Paper29} devote at least one user to secure other users' communication, 
they are not practical  for single user scenarios.  Furthermore, similar to  the works in \cite{Paper12}, \cite{Paper15}, \cite{Paper16}, \cite{Paper20}, and  \cite{Paper26}, the proposed methods in \cite{Paper30} and \cite{Paper31} are restricted to  specific system models since \cite{Paper31} employs an RF tag to back scatter information, which is not available in all the networks, and both works assume  that the legitimate nodes are equipped with multi-array antennas. In addition, the  author in \cite{Paper30} proposed their solution for a single user scenarios and the extension of the method to multi-user  scenarios is  not covered.

In summary,  anti-jamming in the practical case of partially  observable environment against  advanced jammers is an understudied topic in the open technical literature. Thus, in this paper, to ensure safe communication channels for the legitimate users and avoid  channel switching, an anti-jamming mechanism is proposed by deceiving reactive jammers in partially observable environments, which is applicable to both multi user and single user scenarios. Moreover,  we consider the problem of selecting the optimal channel that can be used to deceive the jammer from several available channels.

\subsection{Contributions} 
The main contribution of this paper lies in the design of  an  anti-jamming solution that can be used to fool a jammer by deceiving it into jamming a specific victim channel to secure safe communication channels between legitimate users and an access point (AP). 
Our approach is designed to mitigate the effects of  Reactive  jammers. An important challenge in deception based anti-jamming  is finding the optimal power and channel allocation.    In order to find the optimal channel and  power allocation, availability of channel gains between the network components is necessary. However,  we consider a partially observable environment in terms of the  channel gains between users and the jammer since the position and signal power level of the jammer are not known. Moreover, we study the cases where  the channel gains between users and the AP are known and unknown.
Since  perfect model of
the environment is unavailable, a model-free RL is employed to  solve the power and channel allocation problem. In model-free RL methods, the optimal policy is learned through the agent's  interaction with the environment\cite{Paper32}. 
Moreover, we propose a successive RL-based method that converges three times faster than  regular  RL methods. Moreover, simulation results show  that the proposed anti-jamming technique outperforms previous anti-jamming methods which conduct frequency hopping regardless of channel quality, and the proposed learning strategies closely approach the TRP delivered by the optimal solution. 

The rest of this paper is organized as follows. Section II presents the system model. 
The convex optimization-based anti-jamming for known channel information and RL based anti-jamming for unknown channel information  are proposed  in section III and IV, respectively.
 Simulation results are provided in Section V, and finally, conclusions are draw in Section VI.

\begin{figure}[tbp]
    \begin{center}
\includegraphics[width=0.45\textwidth]{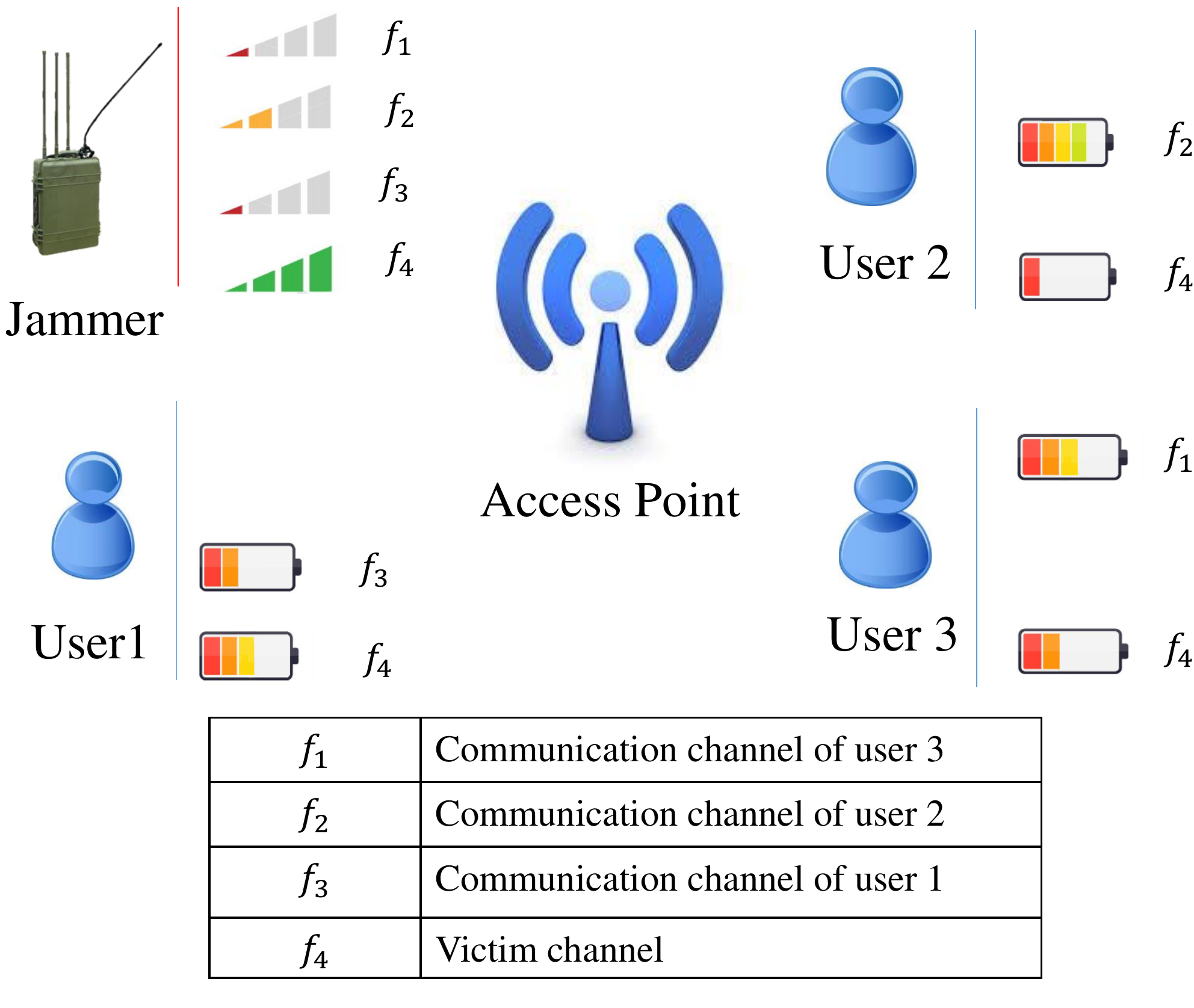}
    \caption{Illustration of our system model.}
    \label{fig1}
    \end{center}
\end{figure}

\section{System Model}
We consider a wireless network consisting of a single AP  that services $N$ users in the presence of a jammer, as shown in Fig. \ref{fig1}. The users and jammer are uniformly distributed in the network area and their positions are fixed. The time is divided into equal slots where, at each time slot, the AP serves the legitimate users using a set $\mathcal{L}$ of $L$ orthogonal channels. We assume that each user can communicate through two channels simultaneously. We assume that the users always have packets to transmit and their transmission power at each time slot is upper bounded by $\bar{P}$. Also,  the jammer's power is limited   but  it is significantly larger than that of the legitimate users. 
All channels between users, the AP, and the jammer are reciprocal and follow a Rayleigh fading model. In addition to the small-scale fading, we consider path loss modeled by $(\frac{ \kappa}{ \kappa_0})^{-\beta}$, where $\kappa$,  $\kappa_0$, and $\beta$ are the distance between nodes, a reference distance, and  the attenuation factor according to the physical environment, respectively. The channel gain between the user $i$ and the AP is  $h_{ci}$, while the channel gain between  user $i$ and the jammer is $h_{ji}$. In what follows, subscripts $c$ and $j$ are used  to denote the AP and jammer, respectively. We consider two distinct scenarios corresponding to the case where the channel gains between the users and the AP are  not known  as well as the case of known channel gains between the users and the AP.

In both scenarios, the channel gains between the users and the jammer are not known, which is the case in practice. Hereinafter, we refer to the availability of  the channel gains between the  users and the  AP as the availability of channel gains. Moreover, we consider a reactive jammer that attempts to disrupt the communication between the AP and users by transmitting its jamming signal over the legitimate users' communication channels. Therefore, we assume that the  data transmitted on the jammed channel is not detected by the AP. Moreover, the users' signals cannot be detected when users interfere with each other over a  given channel. The considered jammer's operation \textcolor{black}{is} detailed next. 

A \emph{reactive jammer}  continuously listens to channels and   jams channels  immediately after sensing an activity \cite{Paper33}. Our considered reactive jammer looks for the channel that has the highest signal power level. It  continuously senses all the channels’ powers and  jams the channel with the highest signal power. In this scheme, if a jammer detects a signal withpower allocationof the   higher power over a given channel while it is jamming another channel, it  instantly switches to the newly detected channel.
  
\section{Convex optimization-based anti-jamming for known channel information}

To address the challenges of securing the communication between legitimate users and an AP against  a reactive jammer   in a partially observable environment, we propose an anti-jamming method that misleads the jammer by using a victim channel. As shown in Fig. \ref{fig1}, engaging the jammer with a specific channel clears other channels for the purpose of secure communications. In this method, every user allocates a specific amount of power to a victim channel to attract the jammer to that channel. To avoid depleting its power, each user has a power consumption limit $ \rho$ for deceiving the jammer. In the multi-user scenario, users can cooperate with each other to select a common victim channel and  announce their actions to other users after each time slot. Moreover, the phase of the users' signals  in the victim channel are assumed to be aligned using the received  jamming signal phases at the users' side.

Here, we study the case in which the jammer jams a single channel in each time-slot, however, the proposed anti-jamming method is applicable to  the case in which multiple channels are compromised by the jammer. In fact, in this scenario, users  absorb the jammer's power in a victim channel to decrease the  jamming power in their communication channels by employing the proposed anti-jamming method.

A major aspect  in the implementation of the proposed method is determining the optimal power allocation of the  victim and communication channels. The optimal resource allocation should achieve the highest achievable TRP  at the AP while deceiving  the jammer with a minimum power consumption in the victim channel. \footnote{Given the fact that considering the sum rate or  TRP as the performance evaluation metrics leads to  the same power and channel allocation, in what follows we focus our study on the TRP.}  The TRP at the AP  excluding  the jammed channel $\bar{G}$, the received signal power at the jammer through the  communication channel $i$ $\hat{F}_{jc_{i}}$, and received signal power at the jammer through the victim channel $\hat{F}_{jv}$ are given in  (\ref{eq1}), (\ref{eq2}), and (\ref{eq3}), respectively. 
 
\begin{equation}\label{eq1}
\begin{aligned}
 & G =\sum_{i=1}^{N}d'^2_i  h_{ci}^2 x_i=\sum_{i=1}^{N}(\bar{P}-d^2_i)  h_{ci}^2 x_i, 
\end{aligned}
\end{equation}
\begin{equation}\label{eq2}
\begin{aligned}
 & \hat{F}_{jc_{i}}= d'^2_i h'^2_{ji},
\end{aligned}
\end{equation}
\begin{equation}\label{eq3}
\begin{aligned}
\hat{F}_{jv}=\left(\sum_{i=1}^{N}d_i  h_{ji}\right)^2\hspace{-0.2cm},
\end{aligned}
\end{equation}
where  $d^2_{i}$ and $d'^2_{i}$ denote the $ith$ user's power allocated 
 
 for deceiving the jammer and communications, respectively, $h_{ji} ^{'}$ is  the channel gain between  $ith$ user's selected channel and the jammer, and $x_i$ is  a  binary  flag that is set to zero if   the $ith$ user's communication channel is jammed or interfered with   other users' communication channels, otherwise it  is set to one.
On the one hand, deceiving the jammer to jam a victim channel is only possible if the jammer always senses the highest signal power  in the victim channel i.e, $\hat{F}_{jc_{i}}\leq \hat{F}_{jv}$ \textcolor{black}{$\forall$  i $\in\mathcal{L}$}. Meanwhile, the users should allocate as much  power as possible  for communication purposes.  Assuming that
  everything  about  the  environment,  including   the  channel  gains  and  jammer  policy  is known, the optimal power allocation problem is formulated as 

\begin{equation}
\begin{aligned}
\label{eq4}
&\hspace{2.5cm} \underset{d_i,d'_i }{\min}{\left(-\sum_{i=1}^{N}{(\bar{P}-d^2_i)}{h^2_{ci}}\right)}, \\
&\hspace{2.5cm}\textnormal{s.t.}  \\
& \boldsymbol{H}\boldsymbol{d}
\geq
\boldsymbol{h}'_j \cdot \boldsymbol{d}'  , \textnormal{ } \boldsymbol{d}\geq{\boldsymbol{\eta}'} 
, \textnormal{ } \boldsymbol{d}\leq {\mathbf{b}},\textnormal{ }
\boldsymbol{d}'\geq {\boldsymbol{\eta}'}
, \textnormal{ } \boldsymbol{d}'  \cdot \boldsymbol{d}'+\boldsymbol{d} \cdot \boldsymbol{d}= {{\boldsymbol{b}}},\\
&\textnormal{where } \\ &\boldsymbol{H}_j = \begin{bmatrix}
h_{j1}&h_{j2}&.&.&h_{jN}\\ 
h_{j1}&h_{j2}&.&.&h_{jN}\\ 
.&.&.&.\\
h_{j1}&h_{j2}&.&.&h_{jN}
\end{bmatrix}\hspace{-0.1cm},\textnormal{ } \boldsymbol{h}'_j= \begin{bmatrix}
h_{j1}^{'}\\ h_{j2}^{'}\\. \\h_{jN}^{'}
\end{bmatrix}\hspace{-0.1cm},\textnormal{ }
\boldsymbol{d} =\begin{bmatrix}
d_1\\ d_2\\ . \\d_N
\end{bmatrix}\hspace{-0.1cm}, \textnormal{ }\\ &
\boldsymbol{d}' =\begin{bmatrix}
d'_1\\ d'_2\\ . \\d'_N
\end{bmatrix}\hspace{-0.1cm},\textnormal{ }
\boldsymbol{\rho}' =\begin{bmatrix}
\sqrt{\rho}\\ \sqrt{\rho}\\ . \\ \sqrt{\rho}
\end{bmatrix}\hspace{-0.1cm}, \textnormal{ }
\boldsymbol{\eta}' =\begin{bmatrix}
0\\ 0\\ . \\ 0
\end{bmatrix}\hspace{-0.1cm}, \textnormal{ }  
\boldsymbol{b} =\begin{bmatrix}
\bar{P}\\ \bar{P}\\ . \\ \bar{P}
\end{bmatrix}\hspace{-0.1cm}.
\end{aligned}
\end{equation}

 To solve (\ref{eq4}), knowledge of  the channel gains between the users and the AP  and the users and the jammer is required. 

Here, in order to characterize the maximum achievable performance of the proposed anti-jamming method, we consider the ideal case in which the channel gains between the users, the AP, and the jammer are known, and we optimally solve problem (\ref{eq4}).

 One can easily verify that (\ref{eq4}) and its feasible set are convex.
Thus, strong duality and the Karush–Kuhn–Tucker (KKT) conditions hold for this problem and the solution can be obtained by applying the KKT conditions on the Lagrangian  of (\ref{eq5}). The dual function of the optimization problem (\ref{eq4}) can be represented as
\vspace{-0.3cm}
\begin{equation}
\begin{aligned}
\label{eq5}
 g(\lambda,\mu) &=\underset{d,d'\in D}{\inf}L_1(d,d',\lambda,\mu)=\underset{d,d'\in D}\inf
\Bigg[ \sum_{i=1}^{N} -(\bar{P}-d^2_i) h^2_{ci} \\ &+ \lambda_{i} \left(-\sum_{k_1=1}^{N}(d_{k_1} h_{jk_1})+ d'_{k_1}h_{jk_1} ^{'}\right)- \lambda_{N+i} d_{i}+\\  & \lambda_{2N+i} (d_{i} -\sqrt{\rho}) -\lambda_{3N+i} d'_{i}+\mu_{i} (d^2_{i}+d'^2_{i}-\bar{P})  \Bigg].
\end{aligned}
\end{equation}

In (\ref{eq5}), only one of the tuple  $(\lambda_i,\lambda_{N+i},\lambda_{2N+i})$ can take a nonzero value, otherwise  $g(\lambda,\mu)$ becomes  infinite.
 
 Applying the KKT conditions on (\ref{eq5}) leads to

1) $\lambda_i (-d_i h_{ji}+d'_1h'_{ji})=0$\textnormal{, which means that} $\lambda_i=0$ \textnormal{or} $d_i h_{ji}=d'_1h'_{ji}$. 

2) $\lambda_{N+i} (-d_i h_{ji})=0$ \textnormal{, and as a result }$\lambda_{N+i}=0$ or $d_i=0\rightarrow d'_i=\bar{P}$.

3) $\lambda_{2N+i} (d_{i}-\sqrt{\rho})=0$\textnormal{, which means that} $\lambda_{N+i}=0$ or $d_i=\sqrt{\rho}\rightarrow d'_i=\bar{P}-\rho$. 

4)  $\lambda_{3N+i} (d'_{i})=0$\textnormal{, which means that} $\lambda_{N+i}=0$ or $d_i=0\rightarrow d_i=\bar{P}$. 

5) $d_i=\frac{h_{ji}\lambda_i+\lambda_{N+i}+\lambda_{2N+i}}{2(\mu_i+ h^2_{ci})}$, and
$d'_i=\frac{-h'_{ji}\lambda_i+\lambda_{3N+i}}{2\mu_i}$.

Many critical points can be obtained by applying the KKT conditions, but only one of them is optimal. The optimal solution is the critical point that has the lowest value of the objective function. 
It is impossible to obtain the solution as a function of the channel gains since their variation affects the KKT conditions. Thus, to assess the proposed method, we use the expectation of the achieved TRP by the AP. The evaluation of this expectation requires the expectation of the channel power gains $h^2_{ci}$,  $i \in\mathcal{X}$ where $\mathcal{X} = \{ i\in \mathcal{N}  | 0\leq i\leq N \}$ and the allocated power for each user’s communication $d'^2_{i}$, $i \in  \mathcal{X} $. The expectation of the channel power gains is known; however, the expectations of the allocated powers are not accessible because the power distribution cannot be  expressed as a function of the channel gains. Therefore, instead of using the solution of the main problem, we adopt the solution of the modified problem that leads to a lower bound to the AP TRP. To this end, instead of considering  the first constraints set $\boldsymbol{H}\boldsymbol{d}
\geq
\boldsymbol{h}'_j \cdot \boldsymbol{d}' $, we assume $   
\boldsymbol{M} 
\boldsymbol{p}\geq
\boldsymbol{b} \cdot  \boldsymbol{h}'_j \cdot \boldsymbol{h}'_j   $, where
\begin{equation}
\begin{aligned}
&\boldsymbol{p}=\begin{bmatrix}
P_1 .  P_2 . . . P_N
\end{bmatrix}^{\intercal},\textnormal{ }P_i = (d_i)^2, \textnormal{ }i \in  \mathcal{X}
\textnormal{, and } \\ 
&\boldsymbol{M}= \begin{bmatrix}
h_{j1}^{2}+h_{j1}^{'2}&.&.&h_{jN}^{2}\\ 
h_{j1}^{2}&h_{j2}^{2}+h_{j2}^{'2}&.&h_{jN}^{2}\\ 
.&.&.\\
h_{j1}^{2}&.&.&h_{jN}^{2}+h_{jN}^{'2}
\end{bmatrix}.
\end{aligned}
\end{equation}
 More precisely, $\boldsymbol{H}\boldsymbol{d}
\geq
\boldsymbol{h}'_j \cdot \boldsymbol{d}'$ can be expanded for each user as 
\begin{equation}
\begin{aligned}
\label{eq7}
h_{j1}d_1+h_{j2}d_2+....+h_{jN}\leq d'_i h'_{ji} \vspace{0.2cm}\textnormal{  for all }  i\in \chi,
\end{aligned}
\end{equation}
\\
and, since $d'_i =\sqrt{\bar{P}-d^2_i}$, (\ref{eq7}) can be presented as 
 \begin{equation}  
\begin{split}
(h_{j1}d_1+h_{j2}d_2+....+h_{jN}d_N)^2\leq (\bar{P}-d^2_i )h'^2_{ji}, 
\end{split}
\label{eq8}
\end{equation}
which shows that (\ref{eq9}) always holds and as a result, $\boldsymbol{H}\boldsymbol{d}
\geq
\boldsymbol{h}'_j \cdot \boldsymbol{d}'$ can be substituted by $ \boldsymbol{M}\boldsymbol{p}\leq  \boldsymbol{b }. \boldsymbol{h'}_{j}.\boldsymbol{h}_{j}$ in (\ref{eq4}).
 \begin{equation}  
\begin{split}
h_{j1}^2d^2_1 +h^2_{j2}d^2_2+....+h^2_{jN}d^2_N\leq (\bar{P}-d^2_i )h'^2_{ji}. 
\end{split}
\label{eq9}
\end{equation}

In this context, a portion of the power received  by the jammer through the victim channel is neglected. Therefore, to achieve a similar signal level as the
main power allocation problem (\ref{eq4}), more power should be consumed in the victim channel and thus, less power remains available for communication purposes. Modifying (\ref{eq4}), we obtain

\begin{equation}
\begin{aligned}
\label{eq10} 
 &\underset{P_i }{\min}{\left(-\sum_{i=1}^{N}(\bar{P}-P_i) h_{ci}^2\right),} \\
&\textnormal{subject} 
 \hspace{0.2cm}\textnormal{to:}
\end{aligned}
\end{equation} 
\begin{align}
&\hspace{-0.6cm}\boldsymbol{M} 
\boldsymbol{p}\geq
\boldsymbol{b} \cdot (\boldsymbol{h}'_j \cdot \boldsymbol{h}'_j), \label{eq11} \\& \hspace{-0.6cm}
  \boldsymbol{p}\geq{\boldsymbol{\eta}'}  
, \label{eq12}\\ & \hspace{-0.6cm}\boldsymbol{p}\leq{\boldsymbol{\rho}' \cdot \boldsymbol{\rho}' }\label{eq13}. 
\end{align}

Applying this modification allows us to obtain the power allocation as a function of  the channel gains. 

The constraints and optimization function in (\ref{eq10}) are linear, and as a result convex. Thus, strong duality and the KKT conditions  hold for this problem too. Since both the optimization function and constraints are linear, the solution is on the border of the feasible set $\mathcal{D}'$\cite{Paper34}, which can be achieved by applying the  KKT conditions on the dual function of (\ref{eq10}).  Therefore, we can find the  power allocation using (\ref{eq11}), a  combination   of (\ref{eq11}) and (\ref{eq12}) or (\ref{eq13}), or (\ref{eq12}) and (\ref{eq13}). In order to study the case where the power allocation is derived from (\ref{eq11}), next, we prove that $\boldsymbol{M}$ is invertible.
 
\begin{proposition}  
\label{pro1}
\textnormal{The matrix $\boldsymbol{M}$, is positive definite  and  as a result invertible.}
\begin{proof}
\textnormal{The proof is provided in Appendix A.}
\end{proof}
\end{proposition}

 From Proposition \ref{pro1}, we can see that $\boldsymbol{M}$ is invertible, thus the power allocation can be derived as
 \begin{equation}\label{eq14}
\boldsymbol {p}=
\boldsymbol{M}^{-1}\boldsymbol{b} \cdot \left(\boldsymbol{h}'_j \cdot \boldsymbol{h}'_j\right).
\end{equation}

Equation (\ref{eq14}) shows that, since the achieved
powers are positive,  the power distribution derived from (\ref{eq11}) is valid for $\boldsymbol{d}\geq{\boldsymbol{\eta}'} $.
Hence,  (\ref{eq11})  is used to obtain the lower bound on the AP TRP obtained from (\ref{eq4}). To find the lower bound, it is necessary to introduce the Sherman–Morrison lemma from [Section 2.7.1]\cite{Paper35}.

 \vspace{0.2cm}

\begin{lemma}
 \textit{If $\boldsymbol{O}$ and $\boldsymbol{O}+\boldsymbol{U}$ are invertible, and $\boldsymbol{U}$ is a rank one matrix, let $g=\textnormal{trace}(\boldsymbol{U}\boldsymbol{O}^{-1}) \hspace{0.15cm}$ and $ g\neq-1$, then
$(\boldsymbol{O}+\boldsymbol{U})^{-1}=\boldsymbol{O}^{-1} -\frac{(\boldsymbol{O}^{-1}\boldsymbol{U}\boldsymbol{O}^{-1})}{1+g}$}.
\end{lemma}

  \vspace{0.2cm}

To use the Sherman–Morrison lemma, first  we represent   matrix $\boldsymbol{M}$ by
 
\begin{equation}\label{eq15}
\boldsymbol{M}= \boldsymbol{H} \cdot \boldsymbol{H} +\boldsymbol{I} \cdot (\boldsymbol{h}'_j  (\boldsymbol{h}'_j)^\intercal),
\end{equation}
where $\boldsymbol{I}$ denotes the identity matrix of  same size as $\boldsymbol{M}$. Making use of the Sherman–Morrison lemma, (\ref{eq11}) can be represented as

\begin{equation}
\label{eq16}
\boldsymbol{p}\leq  \boldsymbol{b} - 
\frac{\bar{P}\begin{bmatrix}
\frac{\sum_{i=1}^{N}h_{ji}^{2}}{h_{j1}^{'2}}, 
\frac{\sum_{i=1}^{N}h_{ji}^{2}}{h_{j2}^{'2}} , 
. . .,
\frac{\sum_{i=1}^{N}h_{ji}^{2}}{h_{jN}^{'2}} 
\end{bmatrix}^{\intercal}}{1+\sum_{i=1}^{N}\frac{h_{ji}^{2}}{h_{ji}^{'2}}}. 
\end{equation}
\vspace{-0.2cm}

\textcolor{black}{The channel gains between nodes result from path loss and Rayleigh fading. Here, $\varkappa $ and $\xi$  are  variables corresponding to  the path loss and Rayleigh fading, respectively.  The Rayleigh fading components of the channel gains between two users at different frequencies are assumed to be independent and identically distributed random variables.} Moreover, since the users and the jammer are uniformly distributed,  $\mathbb{E}(\varkappa_j)$ and  $\mathbb{E}(\varkappa_c)$ are     equal  between all the users and the jammer, and  all the users and the AP, respectively.

Thus, for a given user $k$, (\ref{eq16}) can be reformulated as

\begin{equation}
{P_k}\leq  \bar{P}\left(1 - 
\frac{
\frac{\sum_{i=1}^{N}\varkappa_{ji}^{2}\xi_{ji}^{2}}{\varkappa_{jk}^{2}\xi_{jk}^{'2}}}{1+\sum_{i=1}^{N}\frac{\varkappa_{ji}^{2}\xi_{ji}^{2}}{\varkappa_{ji}^{2}\xi_{ji}^{'2}}}\right).
\end{equation}

From (\ref{eq10}), we can see that the channel selection affects the AP's TRP. In order to find the optimal solution, the communication channel for every user and victim channel should be selected among the $L$ frequency channels. The best channel for deceiving the jammer is the channel that has the highest summation of users' channel power gains, i.e. $\sum_{i=1}^{N}h_{ji}^2$.  The communication channel selection can be conducted by two methods. First, by choosing channels with the lowest gains between users and the jammer to mitigate TRP at the jammer side and second, by selecting channels with the highest gains between the users and AP to increase the TRP at the AP. Intuitively, the second approach  is most likely the   optimal one, however in some cases, selecting the communication channels based on the lowest channel gains between the users and  jammer obtains a higher performance. Thus,  we will consider both  cases. Hereinafter, we name these two approaches \textit{APP1} and \textit{APP2}, respectively.  

\vspace{0.5cm}
The channel power gains of the different frequencies and users are independent,  and as a result the summation of channel power gains from $N$ users over different frequency channels are also independent. Given to the fact that $\underset{\varkappa_{ji},\xi_{ji}}{\mathbb{E}}(\max (\sum_{i=1}^{N}h_{ji}^{2}))\geq\underset{\xi_{ji}}{ \mathbb{E}}(\max(\underset{\varkappa_{ji}}{\mathbb{E}}(\sum_{i=1}^{N}h_{ji}^{2}))$ where $\underset{\xi_{ji}}{ \mathbb{E}}(\max(\underset{\varkappa_{ji}}{\mathbb{E}}(\sum_{i=1}^{N}h_{ji}^{2}))=\mathbb{E}(\varkappa_{j})\mathbb{E}(\max(\sum_{i=1}^{N}\xi_{ji}^{2}))$ and  $i \in\mathcal{X}$,  $\mathbb{E}(P_k)$ can be represented as 
 
\begin{equation}\label{eq18}
\mathbb{E}(P_k)\leq  \bar{P}\left (1 - \mathbb{E}(
\frac{
\frac{\sum_{i=1}^{N}\xi_{ji}^{2}}{\xi_{jk}^{'2}}}{1+\sum_{i=1}^{N}\frac{\xi_{ji}^{'2}}{\xi_{ji}^{2}}})\right).
\end{equation}

The distribution of $\max \left(\sum_{i=1}^{N}\xi_{ji}^{2}\right)$ among $L$ available channels will be

\begin{equation}\label{eq19}
\begin{aligned}
f_{\textnormal{max}}(Z,N,\lambda,L)&=L\left(\frac{ \lambda^N Z^{(N-1)}  e^{-\lambda Z}   }{ (N-1)!}\right)\times\\&\left(1-\sum_{i=0}^{N-1}\frac{  e^{-\lambda Z}(\lambda Z)^i  }{ i!}\right)^{L-1}\hspace{-0.6cm},
\end{aligned}
\end{equation}
The expectation of $\max \left(\sum_{i=1}^{N}\xi_{ji}^{2}\right)$ does not  have a closed form; however, it can be calculated numerically. In what follows, for  notational convenience, the $E(\max \sum_{i=1}^{N}\xi_{ji}^{2})$ is denoted by  $\Gamma$.

 \textit{APP1}  helps users  consume less power for deceiving the jammer. 
In this scheme, a communication channel is assigned to each user that has the lowest channel gain $h''_j$ between the user  and  jammer. Thus, the expectation of $\mathbb{E}(h_{j}^{'2})$ is equal to $\mathbb{E}(\min(h_{j}^{'2})=h_{j}^{''2})$ at the available channels for each user and  $\mathbb{E}(P_i)_{\textnormal{\textit{APP1}}}$ can be derived  as next.
\textcolor{black}{
\begin{proposition}  
\label{pros2}
Using \textit{APP1} as a communication channel selection method among $L$ available channels, \textnormal{$\mathbb{E}(P_i)$,  $\forall$  $i \in\mathcal{X}$} is given by
\begin{equation}\label{eq20}
\mathbb{E}(P_i)_{\textnormal{\textit{APP1}}}\leq \frac{\bar{P}\sum_{k_1=0}^{N-1}\frac{1}{\lambda N (N-k_1)(L-1-k_1)}}{\sum_{k_1=0}^{N-1}\frac{1}{\lambda N (N-k_1)(L-1-k_1)}+\Gamma}.
\end{equation}
\begin{proof}
\textnormal{The proof is provided in Appendix B.}
\end{proof}
\end{proposition}}

Since the channel allocation in \textit{\textit{APP1}} is performed based on the channel gains between users and the jammer regardless of the channel gains between the users and the AP, the expectation of $h_{ci}^2$ is equal to $ \frac{\mathbb{E}(\varkappa^2_{ci})}{\lambda}$. Therefore, based on Proposition 2,  the expectation of the total received signal power (ETRP) at the AP can be presented as
 
\begin{equation}
\label{eq21}
\begin{aligned}
&C_1=\mathbb{E}\left( \sum_{i=1}^{N}   (\bar{P}-P_i) h_{ci}^2\right) = \sum_{i=1}^{N}\left(\bar{P}-P_i \right)  \mathbb{E}\left(h_{ci}^2\right)\\ &=N\bar{P} \left(\frac{\Gamma}{\sum_{k_1=0}^{N-1}\frac{1}{\lambda N(N-k_1)(L-1-k_1)}+\Gamma}\right)\left(\frac{\mathbb{E}(\varkappa^2_{ci})}{\lambda}\right).
\end{aligned}
\end{equation}

\textit{APP2} focuses on enhancing the AP's TRP by selecting the channel with the highest gain among the available channels between the users and the AP. The constraints  of (\ref{eq10})  are independent from the channel gains between the users and the AP. Thus, from (\ref{eq16}), we can easily write
 
\begin{equation}\label{eq22}
\mathbb{E}(P_i)_{\textnormal{\textit{APP2}}}\leq\frac{\bar{P} }{1+\Gamma}.
\end{equation}
 According to the \textit{APP2} policy, the ETRP of the AP will be
 
\begin{equation}\label{eq23}
\begin{split}
C_{\textrm{AP}}=\sum_{i=1}^{N} {\left(\bar{P}-\mathbb{E}(P_i)\right)}      \mathbb{E}\left( \max \left(h_{cil}^2,  l\in(1,...,L-1)\right)\right),
\end{split}
\end{equation}
where $\bar{P}-\mathbb{E}(P_i)=\frac{\bar{P} }{1+\Gamma}$ and $\mathbb{E}\left( \max \left(h_{cil}^2,  l\in(1,...,L-1)\right)\right)$ is the expectation of a random variable  resulting from the selection of the maximum value among $L-1$ random variables, which  random variables are the channel power gains between the users and the AP. The following proposition derives \textcolor{black}{$\mathbb{E}\left( \max \left(h_{cil}^2,  l\in(1,...,L-1)\right)\right)$}.

\begin{proposition}  
\textnormal{The expectation of $\max \left(h_{cil}^2, l\in(1,...,L_1)\right)$ over $L_1$  number of channels and $N$ number of users  when  \textit{APP2} policy is used for the channel allocation is}
\label{pros3}
\begin{equation}\label{eq24}
\begin{aligned}
&\mathbb{E} \left({\max \left(h_{cil}^2, l\in(1,...,L_1)\right)}\right)=\mathbb{E}(\frac{\varkappa^2_{ci}}{\lambda}) V(L_1,N)=\\ & \frac{\mathbb{E}(\varkappa^2_{ci})}{\lambda N}\bigg(\sum_{\underset{ -\forall (k_1,...,k_N)=0}{k_1,...,k_N \in[0, N]} }
 {{L_1}\choose{k_1}}...{{L_1}\choose{k_N}}\frac{(-1)^{(1+k_1+...k_N)}}{( k_1+...+k_N)}+\\& \sum_{\underset{-\forall( k_1,...,k_{N-1})=0}{k_1,...,k_{N-1}\in[0, N] }} 
 {{L_1-1}\choose{k_1}}...{{L_1-1}\choose{k_{N-1}}}\frac{(-1)^{(1+k_1+...k_{N-1})}}{( k_1+...+k_{N-1})}\\&+...+\sum_{k_1=0}^{L_1+1-N} 
 {{L_1+1-N}\choose{k_1}}\frac{(-1)^{(k_1+1)}}{ k_1}\bigg).
 \end{aligned}
\end{equation}
 
\begin{proof}
\textnormal{The proof is provided in Appendix C.}
\end{proof}
\end{proposition}
\vspace{-0.2cm}
Using the result of \textcolor{black}{ Proposition \ref{pros3}} in (\ref{eq23}) leads to
 
\begin{equation}\label{eq25}
C_2=\frac{ \Gamma  \bar{P} }{(1+\Gamma )}NV(L-1,N)\mathbb{E}(\varkappa^2_{ci}).
\end{equation}

\textcolor{black}{To evaluate the  performance of the proposed method, we compare  the obtained ETRP with the expectation of the maximum achievable TRP.
The maximum achievable TRP at the AP   is obtained by allocating the channel with the highest gain to each user in the absence of jammers, which  for N
users and L channels can be calculated as (\ref{eq26})}
 
\begin{equation}\label{eq26}
\begin{split}
\Upsilon_{\textnormal{Top}}=\bar{P} \sum_{i=1}^{N} \hspace{0.1cm}   {\max(h_{cil}^2,  l\textcolor{black}{\in\mathcal{L})}}.
\end{split}
\end{equation}

Since the channel gains are randomly distributed, the maximum achievable TRP of the AP changes according to the channel gains' variations. Thus, we take the expectation of  $\Upsilon_{\textrm{Top}}$.
 
 \begin{equation}\label{eq27}
\begin{aligned}
C_{\textrm{Top}}&=\bar{P} \sum_{i=1}^{N} \hspace{0.1cm} \mathbb{E}(  {\max(h_{cil}^2, l\textcolor{black}{\in\mathcal{L})})}).
\end{aligned}
\end{equation}
 
The expectation of  the maximum achievable TRP can be obtained using the result of Proposition \ref{pros3}.
The TRP rates of  the proposed method for both \textit{APP1} and \textit{APP2}  are given in (\ref{eq28}) and (\ref{eq29}) respectively. 
 
 \begin{equation}\label{eq28}
\begin{split}
\frac{C_{1}}{C_{\textrm{Top}}}=\frac{\frac{\Gamma}{\sum_{j=0}^{N-1}\frac{1}{ \lambda(N-j)(L-1-j)}+\Gamma}}{V(L,N)},
\end{split}
\end{equation}
 
 \begin{equation}\label{eq29}
\begin{split}
\frac{C_{2}}{C_{\textrm{Top}}}=\frac{\Gamma V(L-1,N)}{(\Gamma+1)V(L,N)}.
\end{split}
\end{equation}

 Since users are uniformly distributed in the network,     $\mathbb{E}(\varkappa_j)$ for all the users are equal, and as a result,  (\ref{eq28}) and (\ref{eq29})  are independent of $\mathbb{E}(\varkappa_j)$. Thus, as long as the users are uniformly distributed in the network and channel gains follow a Rayleigh fading model,   the expectation of the performance of  the proposed method is higher than the obtained lower bound, regardless of the jammer location.  

In our considered model, legitimate users need to find the optimal victim channel and power allocation to gain the highest TRP  while the highest  signal level is sensed in the victim channel at the jammer side. In order to find the optimum victim  channel  and power allocation, knowledge of all the  channel gains between the users, AP, and jammer  are needed. However, in realistic scenarios, the  channel gains between the users and the jammer are not known. Moreover, in some cases, channel gains between users and the AP are hard to detect  due to destruction of feedback links by the jammer or the lack of feedback links. Thus, it is necessary to adopt a method that finds the channel selection and power allocation without knowledge of the channel gains.

\section{Reinforcement learning based anti-jamming for unknown channel information}
In the considered system model, the TRP of the users only depends on the power and channel allocation, and  the   channel that is jammed by the jammer at each slot. Thus, the TRP  follows the Markov property and  the interaction between the users and  jammer can be formulated as a Markov decision process (MDP). However, in this  context, transition probabilities cannot be predicted   due to the dynamical   environment and lack of prior knowledge
about the channel gains. Thus,
a model free  RL algorithm approach  is employed to solve the MDP  with unknown   transition probabilities \cite{Paper36}.  
In this context, users select the channels and the corresponding power allocation, and  receive rewards according to the received TRP and success in deceiving the jammer. Thus, as a result of the interaction with the environment within an RL structure, users can find the optimal channel and sub-optimal power allocation.  Next, we explain the elements of the two proposed  RL approaches.

\subsection{Reinforcement learning elements}
Our considered tabular RL method  is defined by a tuple
$<\mathcal {S},\mathcal {A}, R(\cdot)>$, where $\mathcal {S}$ represents the state space,  $\mathcal {A}$ is the action space,
and $R(\cdot)$ is the immediate reward of the system.  As mentioned earlier, for unknown channel gains between the users and the jammer, we consider the scenarios where the channel gains between the users and the AP  are not available,  as well as the case where the channel gains between the users and the AP are available. 

In the first scenario, the state set $\mathcal {S}$ includes all possible combinations of the victim and communication channels of every user while the action set $\mathcal {A}$ includes different combinations of the two channels that each user can select for deceiving the jammer  and communication purposes with the allocated power for the victim channel  taken from the set  $[0 , \rho]$. Moreover, due to the fact that in tabular RL methods, states and actions are discrete spaces, continuous variables that are included in the states or actions set should be quantized. Thus, in our work, we quantize the power with a quantization step of $\frac{\rho}{ \chi}$, where $\chi$ is the number of samples among the $[0, \rho]$. Precisely,  the states and actions sets can be presented as $\mathcal {S}$ =$ \{\boldsymbol{s}_{c1},...,\boldsymbol{s}_{cN}, \boldsymbol{s}_v\} $ and   $\mathcal {A}$ = $\{ \boldsymbol{a}_{c1},..., \boldsymbol{a}_{cN}, \boldsymbol{a}_v, \boldsymbol{a}_{P_1},..., \boldsymbol{a}_{P_N}\} $, respectively, where  $\boldsymbol{s}_v$ and $\boldsymbol{s}_{cN}$ denote the state corresponding to the selected victim and communication channels, $\boldsymbol{a}_{ci}$  corresponds to the communication channel of user $i$ (\textcolor{black}{ $i \in\mathcal{X}$})  among $L$ channels, $\boldsymbol{a}_v$ corresponds to the victim channel action selection among $L$ channels, and $\boldsymbol{a}_{P_{i}}$ corresponds to the  victim channel power among $\chi+1$ power steps for user $i$ (\textcolor{black}{ $i \in\mathcal{X}$}). The size of the state and action sets are $L^{N+1}$ and $L^{N+1} (\chi+1)^{N}$, respectively. 

In the second scenario, the power allocation and  victim channel must be determined by the RL. Thus, the state and action sets of the considered RL   are $\mathcal {S}$ = $\{\boldsymbol{s}_v\} $ and $\mathcal {A}$ = $ \{\boldsymbol{a}_v , \boldsymbol{a}_{P_1},..., \boldsymbol{a}_{P_N} \}$ respectively, where  $\boldsymbol{s}_v $ denotes the state of the selected victim channel, $\boldsymbol{a}_v $ corresponds to the victim channel action selection among $L$ channels, and $\boldsymbol{a}_{P_{i}} $ corresponds to  the victim channel power among $\chi+1$ power steps  for user $i $ (\textcolor{black}{ $i \in\mathcal{X}$}). Due to availibilty of the channel gain between the users and AP, the size of the state and action sets are reduced to $L$ and $ L(\chi+1)^{N}$, respectively.

For
both scenarios, we use the following function to reward the user's channel selection of  the power distribution  and power allocation
\begin{equation}
\label{eq30}
 R(\boldsymbol{d}, \boldsymbol{d'}, \boldsymbol{w},  \zeta)={{\frac{ {G w_1}}{\bar{P}} }}-{{\zeta w_2}}-{{w_3\sum_{i=1}^{N}d^2_i  }},
\end{equation}

\noindent where $w_i$ is the considered weight for  element $i$ of the reward function and $\zeta$ is a binary flag indicating whether the selected victim channel is jammed or not. The reward function consists of three elements, each  defined to make agents follow a  specific behavior. The first term $\frac{G}{\bar{P}}$ encourages users to discover communication channels and power distributions that lead to the highest possible TRP at the AP. The second term $-\zeta w_2$ evaluates the \textcolor{black}{victim channel  and power allocation  action selection by checking whether the victim channel is jammed ($\zeta = 0$) or not ($\zeta = 1$),  and in case that  the victim channel is not jammed,   penalizes the agents by $-w_2$}. The third term $w_3\sum_{i=1}^{N}d^2_i $ is subtracted from the action reward to penalize agents for consuming power excessively in the victim channel.  In what follows, we propose two RL techniques to find the optimal anti-jamming policy for the two previously mentioned scenarios.

\subsection{Anti-jamming without channel information}
\textcolor{black}{Due to lack of channel information,
the behavior of the jammer is not predictable, and thus state transition probabilities  are not available. Among the RL techniques, Monte Carlo and temporal difference (TD) methods do not depend on the state transition probabilities and can learn directly from  visiting the environment\cite{Paper37}. The Monte Carlo method is  not applicable for continuous tasks since the value of a state  is determined at the  end of the episode.  TD learning method is practical for continuous tasks since   the state value is obtained without  waiting for a final outcome.
Q-learning is one of the commonly used  TD methods. }
According to (\ref{eq4}), the AP's TRP  is a function of the selected channels and the corresponding power allocations.
The state-action pair structure of the Q-learning is suitable  for our problem of channel allocation  since the selected channels can be considered as the state and the joint channel selection,  and power allocation can be considered as the users' action. Thus, we employ tabular Q-learning to find the sub-optimal power and optimal channel allocation.

We propose Algorithm 1, in which the Q-learning
method is employed to obtain the power and channel allocation \textcolor{black}{for the  first scenario}. In the Q-learning method, the value of each action $A_t$ and state $S_t$ pair at time slot $t$, $\boldsymbol{Q}(S_t,A_t)$, is determined by visiting different environment states and estimating the value of the  corresponding upcoming states. In tabular Q-learning method, the estimation accuracy is increased by visiting states during the exploration phase and substituting the main value of each state-action pair using the so-called Bellman update rule as follows
\begin{equation}\label{eq31}
\boldsymbol{Q}(S_t,A_t) \leftarrow \boldsymbol{Q}(S_t,A_t)+
\alpha [R+\gamma \underset{a}{\textnormal{max}} \boldsymbol{Q}(S_{t+1},a)-\boldsymbol{Q}(S_t,A_t)],
\end{equation}
where $\alpha$ and $\gamma$ represent the learning rate and discount factor. 
In the considered problem, the states and actions  are defined according to the channel and power allocation of each user. 

The action selection at each state is performed based on the $\epsilon $-greedy policy. In this work, $\epsilon $ is set to one for primitive iterations and then, it is gradually decreased to near zero.  
 
\begin{algorithm}[t ]
\scriptsize
\label{al1}
Algorithm parameters: $\chi$, $\alpha =0.9$, $\gamma =0.9$, $\epsilon = 1 $, $\epsilon_{\textrm{thr}}  $, $\epsilon_J = 0.1 $\;

Initialize $ \boldsymbol{Q}_i(s, a)$ for each user, for all $s \in \mathcal{S}, a \in \mathcal{A}(s)$,  $ \boldsymbol{Q}_i(\cdot, \cdot) = 0$,\\
$k=0$, $k_1=0$, $\Phi_{\epsilon}$ and $\Pi_{\textnormal{Iteration}}$\;
\While{\textnormal{k} $\leq$ $\Pi_{\textnormal{Iteration}}$}{
    \ForEach{\textnormal{step of episode}}{

        z$\leftarrow$Rand([0, 1]);

  \eIf{z$\leq \epsilon $}{
Random stream producer selects a channel randomly as the victim channel;

 \For{i=1:N}{
User $i$ chooses its action randomly\;
}
   }{
 \For{i=1:N}{
User  $i$ chooses its action using greedy policy\;  
} 
  }
        Jammer selects its channel based on its policy\;
        observe $R$, $S'$\;
        Each user updates its Q-table;

        $ \boldsymbol{Q}_i(S_i, A_i) \leftarrow (1-\alpha) \boldsymbol{Q}_i(S_i, A_i) + \alpha [R + \gamma \underset{a}{\max}  \boldsymbol{Q}_i(S_i', a)]$\;
        $S_i \leftarrow S_i'$\;

     $\epsilon \leftarrow  \max(\exp(-\frac{k}{\Phi_{\epsilon}}),\epsilon_{\textrm{thr}})$\;
    }
  Update $P_{\textnormal{i}}^k, i\in\mathcal{N} $\; 
 \eIf{$P_{i}^k =P_{i}^{k-1},\forall i\in\mathcal{N} $}{
  $ k_1\leftarrow k_1+1$\;
   }{
  $ k_1\leftarrow 0$\;
  }
\If{$k_1 =\Phi_{\epsilon} $}{
Break
}
k $\leftarrow $ k+1,
}
\caption{Proposed Q-learning for Anti-Jamming}
\end{algorithm}

\begin{figure}[t]
\centering
\includegraphics[width= 0.5\textwidth]{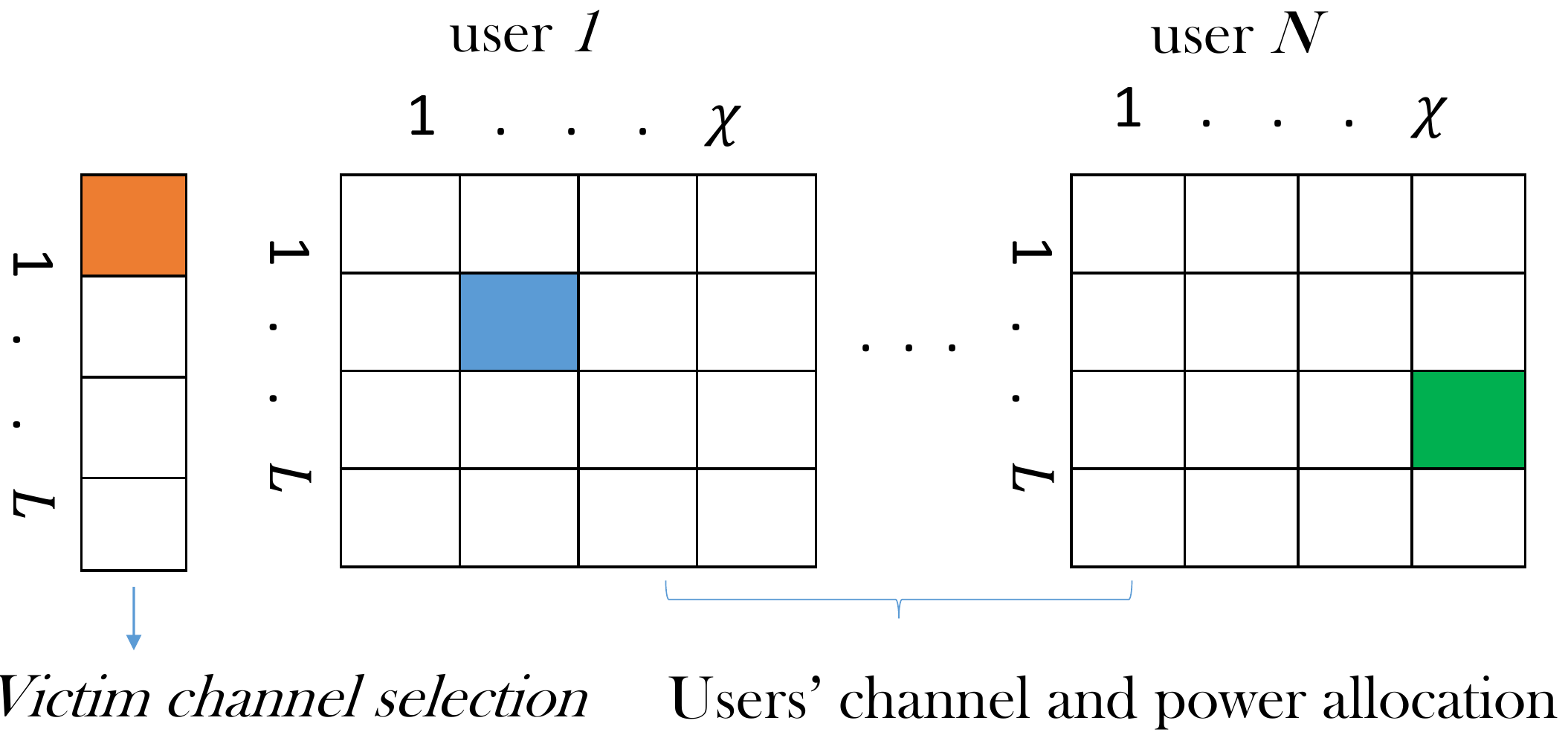}
\caption{Random channel selection in the proposed distributed learning scheme. }
\label{fig2}
\end{figure}

In order to find the optimal solution, we employ distributed Q-learning. In the adopted method, each user keeps a Q-table that includes the possible states and actions of all users. In the considered learning strategy, all the users just follow a random stream for their action selection mode, which results in the same schedule of greedy and random actions. This random stream can be produced by any of the users and announced to others. Moreover, in the random action selection mode, users' actions are not selected based on their joint actions taken from Q-table cells. In fact, as shown in Fig. \ref{fig2}, each user selects the action from its own available actions among $L (\chi+1)$ actions regardless of other users' actions. The victim channel is  selected by the user that is chosen to produce a random stream and other users follow its step. According to this policy and the fact that users are rewarded equally, the users' Q-tables are identical.

The consistency of the Q-tables  allows   users to select a joint action that benefits all of them and prevents interference in greedy action mode. 
Precisely, a substantial  portion of the users' reward depends on their  obtained normalized  $G$, where, according to the selected exploration policy ($\epsilon$-greedy), each user attempts to take an action that  returns a higher reward.  Thus, in order for users to get a higher reward, they should take an action  that returns a higher normalized  $G$. Furthermore,
  users are not allowed to allocate  more power than $\rho$ in the victim channel. Therefore, users cannot devote themselves or other users to gain a higher reward. Thus, in each time slot, the action that leads to a higher TRP  and considers all the users' satisfaction is taken.  The proposed method is detailed in Algorithm 1.

\subsection{ Anti-jamming when  channel gains between users and the AP are known}
Next, we consider that the channel gains between the users and the AP are available. Here, given the fact that channel gains between users and the jammer are unknown, we employ RL. In addition, inspired by the Bisection search method, we propose the successive reinforcement learning (SRL) to enhance the convergence speed. In this method, instead of exploring the environment with a high resolution, we approach to the optimal solution by increasing the exploration resolution gradually. Precisely, instead of deriving a tabular RL with a table including numerous states and actions, successive tabular RLs with small tables are employed. Thus, in the exploration with a low resolution, a significant number of  states and actions that return  low rewards are filtered and  exploration with a high  resolution is performed around the state-action  pair that  returns the highest rewards.

\begin{algorithm} [ht!]
\label{al2}
\scriptsize
Algorithm parameters: $\epsilon_J = 0.1 $, $\Omega_{\textrm{Q}}=\frac{\rho}{\chi_{\textrm{Q}}}$, $\alpha =0.9$, $\gamma =0.9$, $\Phi_{\epsilon}$, \\$\Pi_{\textnormal{Iteration}}$, $\Psi_{\textnormal{end}}$, $\Omega_{\textrm{TD}}=\frac{\tau}{\chi_{\textrm{TD}}}$, and $ \boldsymbol{P}_{\textrm{TD}}=[-{\tau}:\frac{2\tau}{\chi_{\textrm{TD}}}: {\tau}]$\;
\vspace{0.1cm}
Q-learning Part\\
\vspace{0.1cm}
Employing Algorithm 1 with the initialized parameters  to obtain \\the primary  power distribution $\boldsymbol{P}_{\textnormal{Offset}}$;\\
\vspace{0.1cm}
TD(0) Value Iteration Part\\
\vspace{0.1cm}
Initialize: $\epsilon = 1 $, $\epsilon_{\textrm{thr}}$,
 $\textnormal{flag}=0$, $\boldsymbol{P}_{\textnormal{ter}} = \boldsymbol{P}_{\textnormal{Offset}}$, $k=0$, and $k_1 = 0$\;
\While{$\textnormal{flag}=0$}{
Initialize $\boldsymbol{V}(s)$, for all $s \in \mathcal{S}^+$,  $\boldsymbol{V}( \cdot) = 0$, and $\boldsymbol{P}_{\textnormal{Offset}} = \boldsymbol{P}_{\textnormal{ter}}$\;
\While{ $k \leq$ $\Pi_{\textnormal{Iteration}}$}{
    \ForEach{\textnormal{step of episode}}{

        z$\leftarrow$Rand ([0, 1]);

  \eIf{z$\leq \epsilon $}{
 \For{i=1:N}{
User $i$ chooses its action randomly from  $1 \textnormal{ to } \chi_{\textrm{TD}}$ power steps;
}
   }{
 \For{i=1:N}{
User $i$ chooses $S_i$  using greedy policy\;  
} 
  }

        Jammer selects its channel based on its policy\;
        observe $R$, $S'_{i}$\; Each user updates its value table\;
               $ \boldsymbol{V}_i(S_i) \leftarrow \boldsymbol{V}(S_i)+
\alpha [R+\gamma \boldsymbol{V}(S'_{i})-\boldsymbol{V}(S_i)]\;$

   $\epsilon \leftarrow  \max(\exp(-\frac{k}{\Phi_{\epsilon}}),\epsilon_{\textrm{thr}})$\;

    }
  Update $P_{\textnormal{i}}^k, \forall i\in\mathcal{N} $\; 
 
  \eIf{$P_{i}^k =P_{i}^{k-1},\forall i\in\mathcal{N}  $}{
  $ k_1 \leftarrow k_1+1$\;
   }{
  $ k_1 \leftarrow 0$\;
  }
\If{$k_1 = \Psi_{\textnormal{end}} $}{
$\boldsymbol{P}_{\textnormal{ter}} \leftarrow \boldsymbol{P}_{\textnormal{Offset}} + [P_{\textnormal{1}},..., P_{\textnormal{N}}]$\;
Break;
}  
 $k\leftarrow k+1$,  
$\tau \leftarrow  \Omega_{\textrm{TD}}$\; 
The new $\chi_{\textrm{TD}}$ is set\;
}
\If{$\boldsymbol{P}_{\textnormal{ter}} = 0 $}{
$\textnormal{flag} \leftarrow  1$\;
}
}
\caption{Successive Reinforcement Learning }
\label{AL2}
\end{algorithm}

In the context of the considered system model, first users employ Q-learning with a power step of $\Omega_{ \textrm{Q}}=\frac{\rho}{\chi_{\textrm{Q}}}$, where $\chi_{\textrm{Q}}$ is the number of samples
in the interval $[0, \rho]$ for the primary power allocation,  to find the primary power allocation and the optimal victim channel. 
 
After obtaining the primary channel and power allocation, to converge with a higher power resolution, the one step temporal difference value iteration method (TD(0)) \cite{Paper36} with a power step of $\Omega_{\textrm{TD} }= \frac{2\tau}{\chi_{ \textrm{TD} }}$ and adjusting power range $[{-\tau}, \hspace{0.1cm} { \tau }]$ is employed, where $\tau $ and $\chi_{\textrm{TD} }$ are the considered power bound and the number of power samples
 for the TD learning, respectively. In the TD(0) value iteration method, an agent follows the exploration policy to explore  the environment states and updates the value of each state as \cite{Paper36}
 
\begin{equation}\label{eq32}
V(S_t) \leftarrow V(S_t)+
\alpha [R+\gamma  V(S_{t+1})-V(S_t)].
\end{equation}

Once more the $\epsilon$-greedy policy is selected for exploration. The TD(0) learning states set includes $\mathcal {S} = \{s_{P_1},...,s_{P_N}\}$, where $s_{P_{i}}$ denotes the state corresponding to the  power, taken from $[{-\tau}:\frac{2\tau}{\chi_{\textrm{TD} }}:{ \tau}]$, that can be added to the primary power of  user $i $ (  $i \in\mathcal{X}$ )  at the victim channel. Moreover, the size of the states set is $(\chi_{\textrm{TD} }+1)^{N}$.

In this scenario, agents adjust  the primary power allocation and receive rewards for the adjustments. The first  implementation of the TD learning is derived  assuming $\tau= \Omega_{\textrm{Q}}$. After the value of each state is determined and the learning process is finalized, the power set that has the highest state value is then added to the primary power distribution and selected as the new power distribution. The learning process is terminated when the state value matrix remains the same over a predefined number of iterations $\Psi_{\textrm{end}}$. The achieved power allocation is fed to the TD learning as the new power offset  to find the new power allocation  while the new power bound $\tau$ is set  to the previous power step $\Omega_{\textrm{TD}}$. This process is repeated until zero power values are determined as the additive power for all the users,  and  a sub-optimal power allocation with  bounded error based on the final quantization step is achieved as stated in Proposition 2. 
 
\begin{proposition} 
\label{pro4}
 \textnormal{ A sub-optimal power allocation  bounded according to the final quantization step  is achieved by SRL algorithm.}
\begin{proof}
\textnormal{The proof is provided in Appendix D.}
\end{proof}
\end{proposition}

The reward function of the TD method is again set to (\ref{eq30}), and since the best victim channel is selected in the primary learning process, the negative reward for penalizing the wrong victim channel becomes zero. In addition, the same strategy  introduced in subsection $\textnormal{IV-B}$, which makes  users keep a similar table, is selected for distributed learning. The full schema of this method is presented in Algorithm 2. 

The SRL approach   reduces the number of   actions significantly, hence, its learning convergence is faster than regular tabular RL methods. Moreover, after the first implementation of SRL, the users obtain a sub-optimal point  and further explorations are done  when a  fairly  high performance is already achieved. In contrast,  in regular Q-learning with the same power resolution, many time slots are needed for the environment  to be explored and most of the exploration is conducted when the obtained TRP is low. Thus, in the same period of  time, SRL can converge to the optimal power allocation with a higher resolution and, as a result, obtains a higher performance. 

After   an adequate number of time slots since the value of  $\epsilon$ decreases to near zero, the state-action pair (or state for TD(0)) that returns the highest reward is  selected. Precisely,  when $\epsilon$ is near zero, an action-state pair  that maximizes (\ref{eq30}) among all the possible action-state pairs is selected.

According to (\ref{eq30}) the state-action that has the highest TRP at the AP and consumes the lowest power for deceiving the jammer is rewarded the most. In the considered learning structures, the  users' power are in the feasible set of (\ref{eq4}) when the victim channel is jammed because in this circumstance,  $\boldsymbol{H}\boldsymbol{d}
\geq
\boldsymbol{h}'_j \cdot \boldsymbol{d}' $   and the other  conditions in (\ref{eq4}) are considered in the users' action ( or state for TD(0)) selection. Hence,  when the victim channel is jammed, the solution of  (\ref{eq4}) maximizes (\ref{eq30}) too since   the solution of $ \underset{d_i,d'_i }{\min}{\left(-\sum_{i=1}^{N}{(\bar{P}-d^2_i)}{h^2_{ci}}\right)}$  and $ \underset{d_i,d'_i }{\max}{\left( {{\frac{ {G w_1}}{\bar{P}} }}-{{w_3\sum_{i=1}^{N}d^2_i  }}\right)}$  at the feasible set of (\ref{eq4}) are the same. Therefore, adopting  the considered reward function and exploration policy  leads to the convergence to the optimal solution. The same rule holds for the SRL. In this scheme,  in the first iteration of the TD method, the power and the victim channel that return  the highest reward are selected, and in the next iteration the resolution of the sampling is increased. At the end,  the best  sub-optimal power allocation  based on the final  quantization power step (according to Proposition \ref{pro4}) and the victim channel  that returns the highest reward  are selected.   
In addition, with a similar power allocation, better channel selection in terms of the channel gain returns a higher reward. Thus, among the different channel allocation possibilities, the one that has the highest summation of channel power gains, i.e. $\sum_{i=1}^{N}h_{ji}^2$  and  channel   gains  (when the channel gains between users and AP are not known) are selected for the victim channel and communication channels, respectively.

In the considered system model, the jammer always attempts to jam a channel that has the highest sensed signal power. Using the proposed anti-jamming method, users provide a victim channel with the highest sensed signal power at the jammer side. Thus, the reactive jammer  prefers to jam the victim channel, and when the  power and channel allocation are optimized, neither the users nor the jammer want to change their situation. 

The proposed   learning methods are based on the model free tabular RL where the computational complexity order of model free tabular RL learning methods is linear as function of number of states and actions $\mathcal{O}(|\mathcal{S}|^2 |\mathcal{ A}|)$ \cite{Paper38}. Thus, in the case  where channel gains between the users and the AP are unknown, the complexity order is   $\mathcal{O}( L^{2N+2} (\chi+1)^{N})$  and for the case in which channel gains are known,it is  $\mathcal{O}(L^{2} (\chi+1)^{N})$. In the  scenario where SRL is employed, the number of power steps ($\chi$) is significantly lower than in regular RL, which remarkably impacts   the convergence speed.

\section{Simulation Results}

In this section, we evaluate our results using extensive simulations. First, we evaluate the performance of the proposed anti-jamming method in terms of the TRP ratio and the necessary power for deceiving the jammer  according to the obtained lower bound. Moreover, we illustrate the variation of the obtained TRP ratios  by solving (\ref{eq4}) as a function of $\rho$. Then,  we compare the obtained TRP ratio with the TRP ratio of  the proposed methods in  \cite{Paper13}, \cite{Paper21}. Besides,  
in order to show that our proposed method outperforms  frequency hopping methods which are conducted regardless of channel quality, we compared the obtained TRP ratio with the TRP ratio of  the  random search channel selection without any jammers. Furthermore, to evaluate the proposed learning strategies,  the obtained TRP ratio  from each learning strategy is  compared with the optimal AP TRP ratio. 
The  ratio is calculated by dividing the TRP of the AP obtained by the aforementioned scenarios to the maximum achievable TRP of the AP without any jammers. In addition, we compare the convergence rate of the proposed SRL with the Q-learning method using the ratio of their obtained TRPs to the optimal TRPs. Finally,
we  show   how much the  proposed  SRL  method  is  successful  in deceiving the jammer to jam the selected victim channel. To this end, we define  a  metric  named  success  rate obtained by calculating the
  ratio that the jammer jams the selected victim channel  over the selection of other channels in implemented trials. Since results are presented as a function of the ratio of the TRP,  power is normalized and thus, we set $\bar{P}=10$  for  each user in each iteration. The power consumption limit for deceiving the jammer is set to $\rho =\frac{\bar{P}}{2}$, the channel power gains are produced by an exponential probability  distribution function with unit mean and variance,  the adjustment weights are set  to $\boldsymbol{W}=[3.5 $ $ 1.5 $ $1.5$ $]$, $\Phi_{\epsilon}=10000$, $\epsilon_{ \textrm{thr} }=0.0001$, and both the learning rate and discount factors are set to $0.9$ for the users and jammer. Statistical results are averaged over a large number of independent runs.

\begin{figure}[t]
\centering
\includegraphics[width= 0.5\textwidth]{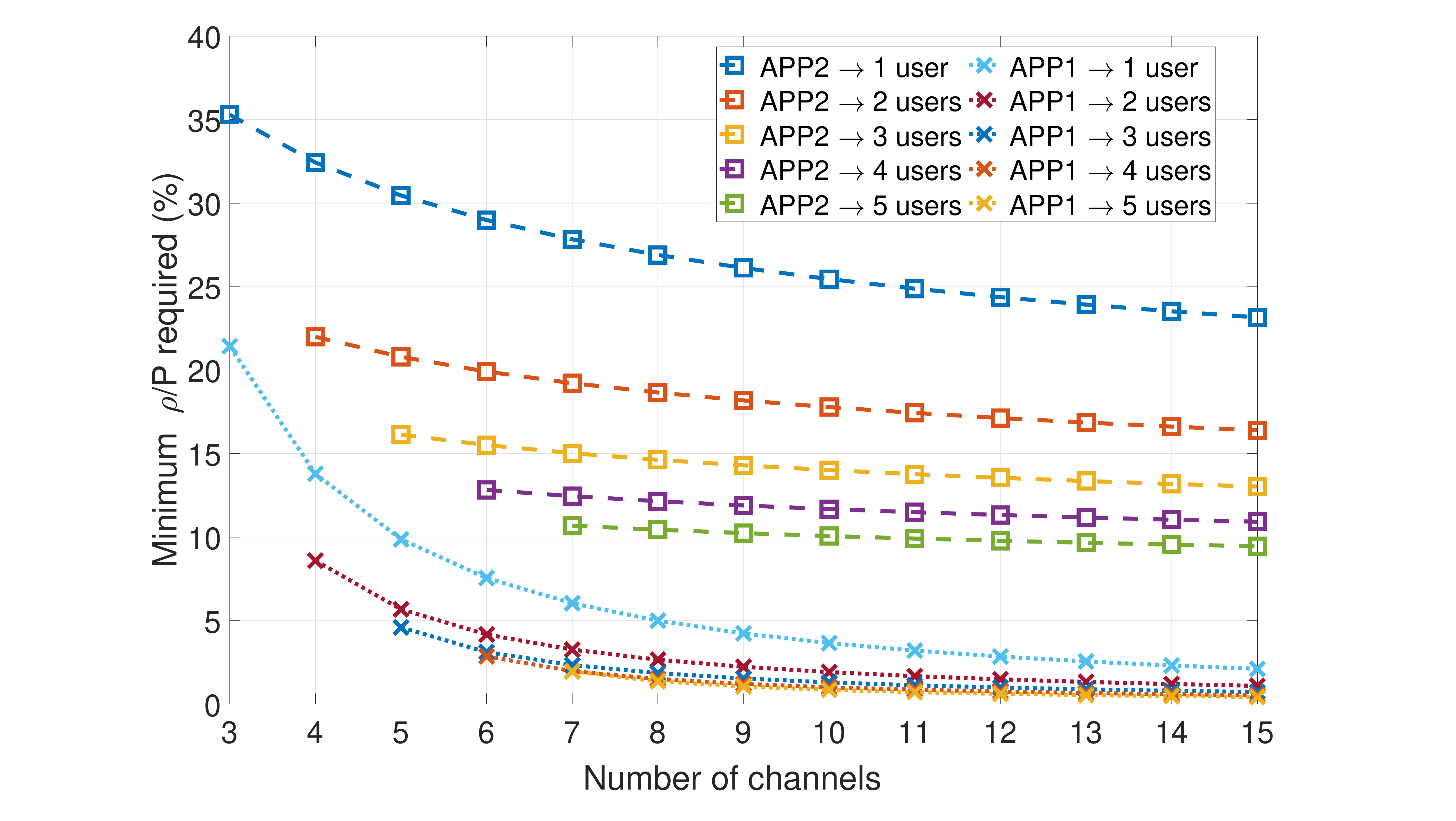}
\caption{Minimum  $\rho$ required for the ratio of total power $\bar{P}$. }
\label{fig3}
\end{figure}

\subsection{Channel selection effects}
The  TRP ratios of the calculated lower bound in section \RN{3} are obtained assuming the power distribution  results from (\ref{eq14}). Equation (\ref{eq16}) shows that the obtained expected powers are valid at  $ \boldsymbol{d}\geq{\boldsymbol{\eta}'} $  since the achieved powers are positive. The third set of constraints is valid if $E(P_i)\leq \rho$ $\textcolor{black}{(i\in\mathcal{X})}$ holds. In Fig. \ref{fig3}, the minimum required power $\rho$ of both channel selection methods according to (\ref{eq20}) and (\ref{eq22}) is shown as a function of  the number of  users, for various system configurations with different numbers of available channels. From Fig. \ref{fig3}, we can see that the highest required power for deceiving the jammer is $35\%$ of the maximum power,  which is the case   when there is one user and four channels. Given to fact that  we assume $\rho=\frac{\bar{P}}{2}$ in our simulations, the power set obtained by the first constraints set satisfies other constraints. 

Fig. \ref{fig3} shows that due to \textit{APP1} policy, with the
same number of users and available channels, the necessary power for deceiving the jammer in  \textit{APP1}  is less than in \textit{APP2}. In both scenarios, with a fixed number of  users, increasing the number of  available channels decreases the required power for deceiving the jammer. The reason behind this is that increasing the number of available channels raises the chance of selecting a victim channel with a higher summation of channel power gains, i.e. $\sum_{i=1}^{N}h_{ji}^2$. The same trend holds for increasing the number of users when the   number of channels  is fixed since  more users allocate power into the victim channel and the jammer can be deceived using less power per user.  For  instance, whenever all  the  channel  gains  between  users  and  the  jammer  are  equal  to  one,  in  a two users scenario, each user has to allocate $\frac{\bar{P}}{5}$ of its power into the victim channel while for a  three users scenario, the necessary power is $\frac{\bar{P}}{10}$.
   
\begin{figure}[!tbp]
\centering
 \includegraphics[width=0.5\textwidth]{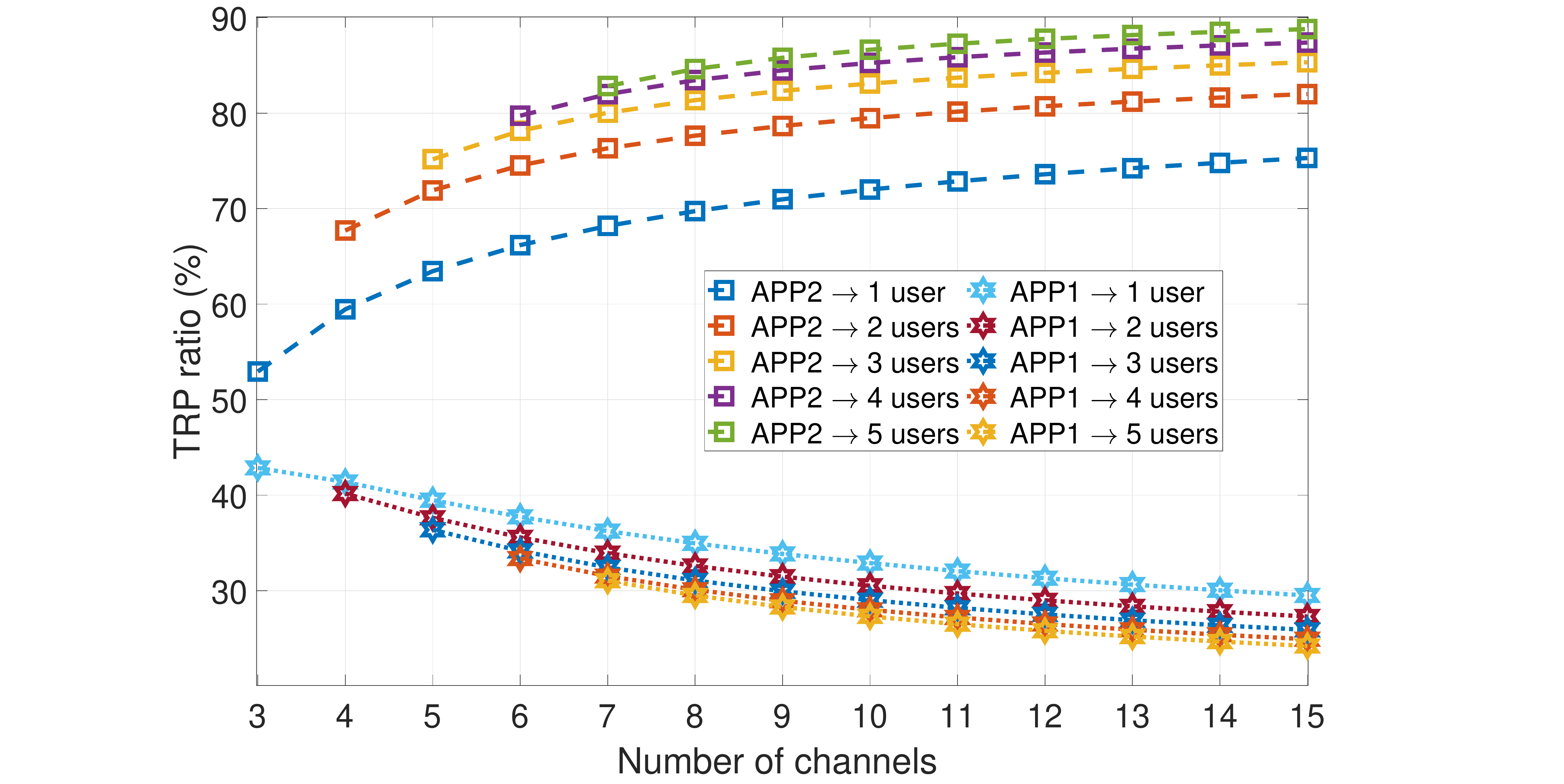}
\caption{  AP's TRP  ratio  obtained from \textit{APP1} and \textit{APP2} approaches. }
\label{fig4}
  \end{figure}

Fig. \ref{fig4} illustrates the ratio of the AP ETRP, obtained from the introduced   channel selection methods, over the expectation of the maximum achievable TRP in the absence of jammers. Our results are shown for one to five users and different available  AP channels.  It is demonstrated  that in \textit{APP2}, the TRP of the AP raises by increasing the number of channels. This growth results from the increase of   $\mathbb{E}(\max(h_{ci}^2,$ $ i\in\mathcal{L}))$ in (\ref{eq24}) by increasing the number of available channels. Moreover, with a fixed number of available channels, increasing the number of users improves the ratio because more users contribute to the allocation of power in the victim channel, and as a result, each user consumes less power for deceiving the jammer. In contrast to  \textit{APP2}, the ratio of \textit{APP1} is reduced by increasing the number of users and channels. The reason for this degradation can be better understood from (\ref{eq21}) and (\ref{eq27}). Equation (\ref{eq21}) shows that the ETRP in \textit{APP1} increases when the number of available channels increases, however, the growth of the expectation of the maximum achievable TRP by increase of the number of available channels (\ref{eq27}) is  more significant than \textit{APP1}. Finally, it is also shown that \textit{APP2}  performs better than \textit{APP1} for all users and channel sizes, and hence we use the ETRPs of \textit{APP2} approach as the lower bound of the main ETRPs  hereinafter.

 Fig. \ref{fig4}  shows that for all the considered number of users, the ETRP growth rate decreases for  any number of channels above ten.  Hence,  it  is  not  necessary to  consider  a  large  portion  of  the  spectrum to select a victim channel. The same trend holds for  increasing the number of users, where the difference between four and five users is negligible.
In addition,  results    show that the  jammer  can be deceived by three users with ten  available channels   with a performance higher than $85 \%$. Therefore, if a large number of users interact with a jammer, allocating the power of   only a few users in the victim channel is enough to mitigate the jamming effect and allow other users to communicate safely without allocating any power into the victim channel.

\begin{figure}[t]
\centering
\includegraphics[width=0.5\textwidth]{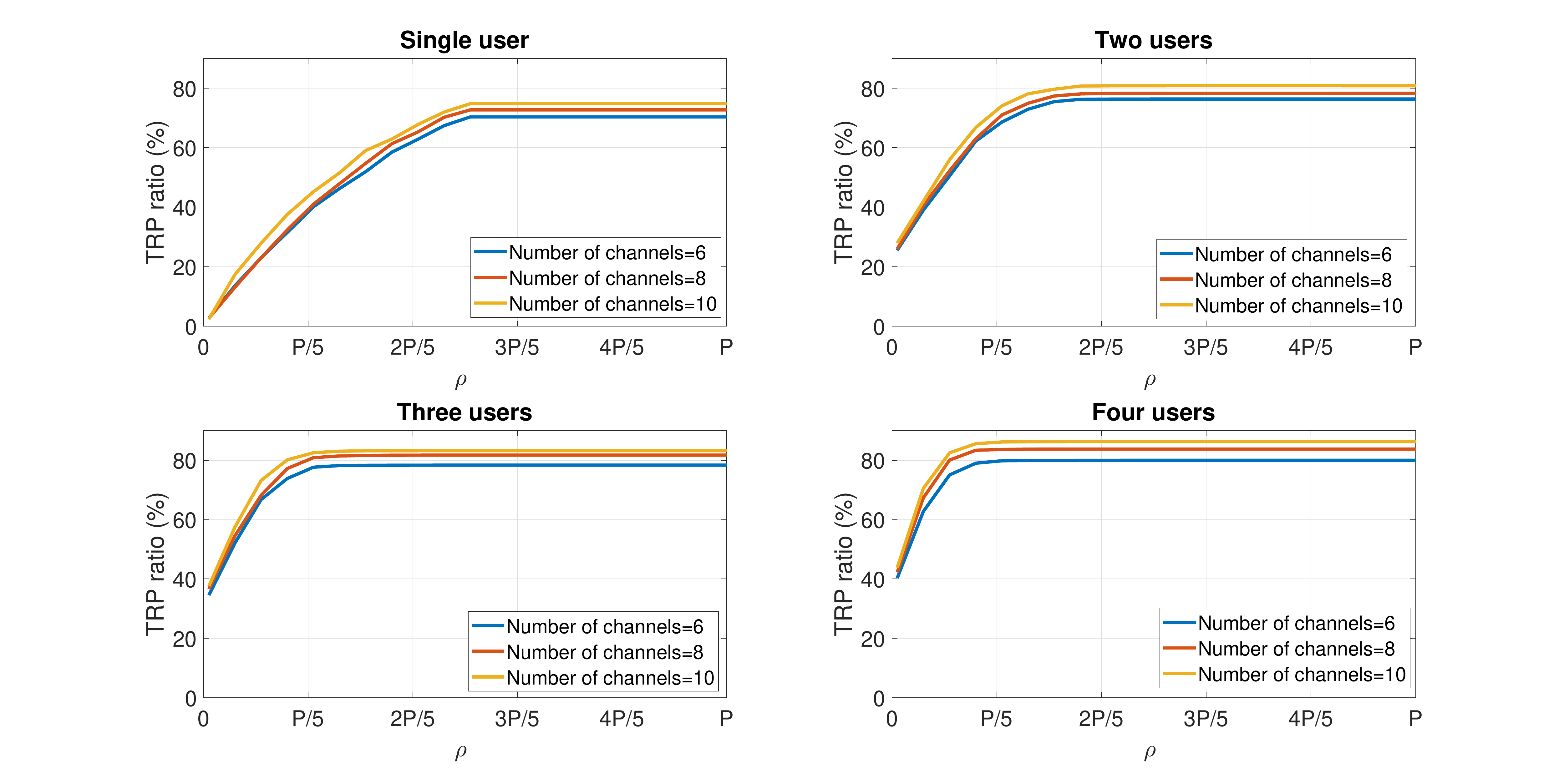}
\caption{TRP ratio   changes as a function of $\rho$. }
\label{fig5}
\end{figure}

In Fig. \ref{fig5}, the average TRP ratio for one to four  users and various numbers of channels  is presented  as a function of $\rho$. Results show that for all the considered channels and users  numbers, increasing $\rho$ increases the TRP up to a certain TRP floor achieved for $\rho$ greater than a threshold.  The reason behind this is that increasing the value of $\rho$ provides opportunity for  the users to allocate the necessary power in the victim channel to deceive the jammer and, as a result, the jammer does not jam the communication channels. Moreover, in the multi-user scenario,
users that have  quality channels to the AP are able to contribute less to the victim channel. In addition, the mentioned threshold is enough for all the users to  deceive the jammer while allowing the users with good  channels  contribute less   in  the victim channel.

\vspace{-0.3cm}
\subsection{Deceiving jammer without channel information}
\subsubsection{Single-user}
\begin{figure}[t]
        \centering
  \includegraphics[width=0.5\textwidth]{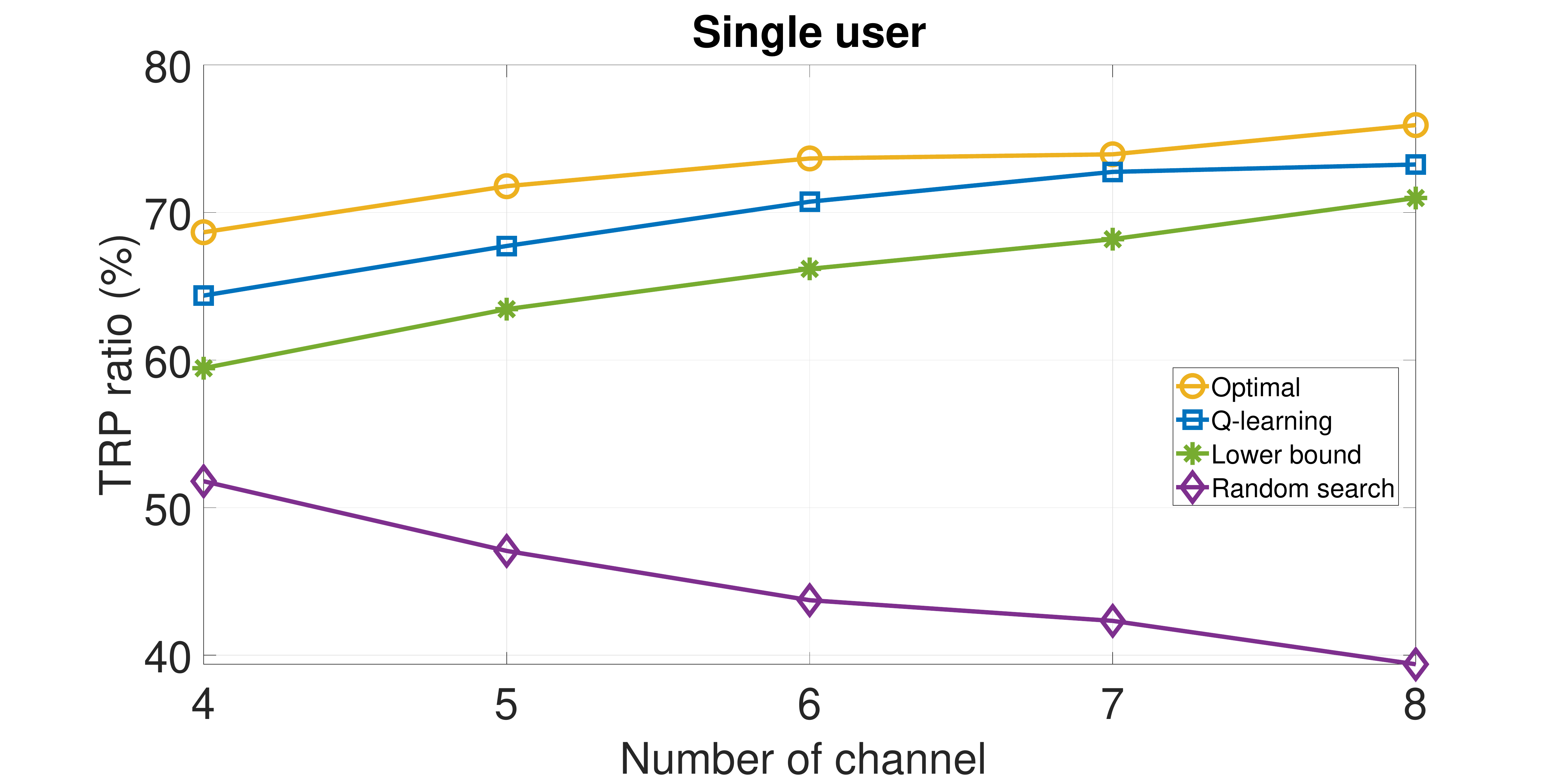} 
       \caption{TRP  for single user scenarios without channel gains.}
        \label{fig6}
\end{figure}

For the single user scheme, the power step is assumed to be $0.2$, and four to eight available channels are considered. 
Fig. \ref{fig6} shows that in the single-user scenario with four available channels, the proposed method can achieve about 60$\%$ of the maximum achievable AP TRP  without having any knowledge of the environment. These results are more promising than the results of the random search in the absence of  jammers, which proves that the proposed method is able to both mitigate the jamming effects and achieve an acceptable TRP. Moreover, the closeness of the average TRP ratio from the optimal solution and Q-learning proves that the success of the adopted learning strategy is not restricted to a specific channel set. 

\subsubsection{Multi-user}
\begin{figure}[t]
        \centering
  \includegraphics[width=0.5\textwidth]{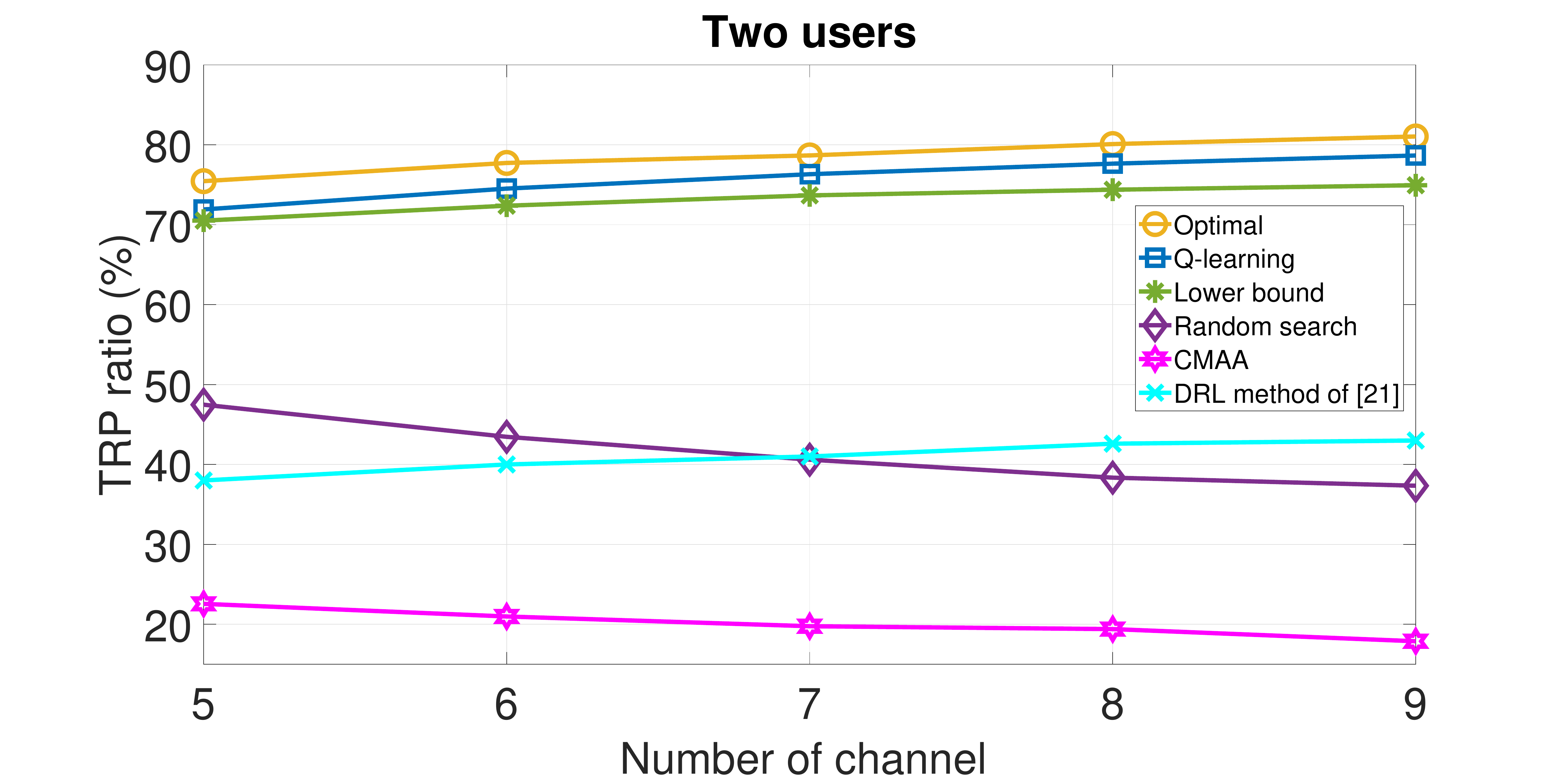} 
        \caption{ TRP  for two users scenarios without channel gains.}
        \label{fig7}
\end{figure}

In the multi-user scenario, users cooperate with each other to deceive the jammer by allocating power in a common victim channel. For this scheme,  we consider two users with five to nine available channels. The power step is set to  two. Fig. \ref{fig7} is similar to Fig. \ref{fig6} but for two users. The AP's TRP ratios obtained by the Q-learning method are higher than the    proposed anti-jamming techniques in \cite{Paper13}, \cite{Paper21}, and random search method for all the considered channel numbers, which shows that the proposed learning strategy is successful in the multi-user case as well. The calculated optimal AP TRP ratio for two users and six channels with full knowledge of the environment shows that the proposed anti-jamming method can achieve a AP's TRP higher than $80\%$ of the maximum achievable TRP. Moreover, the fact that the empirical results are quite similar to the optimal results proves that a near  optimal performance is achievable with the proposed learning strategy. The comparison between the AP TRP ratio of the two-user scenario and the one-user scenario demonstrates that increasing the number of users enhances the  AP's TRP ratio. Compared to the one-user scenario, gaps between the empirical and optimal results in the two-user scenario are higher. The reason for this disparity is that in the two-user model, $P_{\textrm{step}}$ is set to $2$ to decrease the number of states and actions, and thus, the obtained power allocation has an  accuracy of 2 which cannot match the optimal solution.

\subsection{Deceiving jammer with channel information}
Here, we assume that the channel gains between the users and the AP are available. Therefore, the best communication channel for each user is clear and just the power allocation and victim channel selection should be determined.  Algorithm \ref{AL2} is utilized to find the optimal victim channel and sub-optimal power allocation.  \textcolor{black}{ To achieve the power allocation with an accuracy of $0.1$ in Algorithm \ref{AL2}, Q-learning is implemented once and  TD(0) learning  twice. The power step of the Q-learning is set to two, while for the TD learning, the power step of the first iteration is set to $0.5$ and the power variation range is limited to $[-2$, $2]$, and in the second iteration, the power step is decreased to 0.1 and the power variation range is limited to $[-0.5$, $0.5]$}. Simulations are performed assuming three users, while five to nine available channels and for the TD learning part  $\Phi_{\epsilon}$  is set to $1000$.

\begin{figure}[t]
        \centering
  \includegraphics[width=0.5\textwidth]{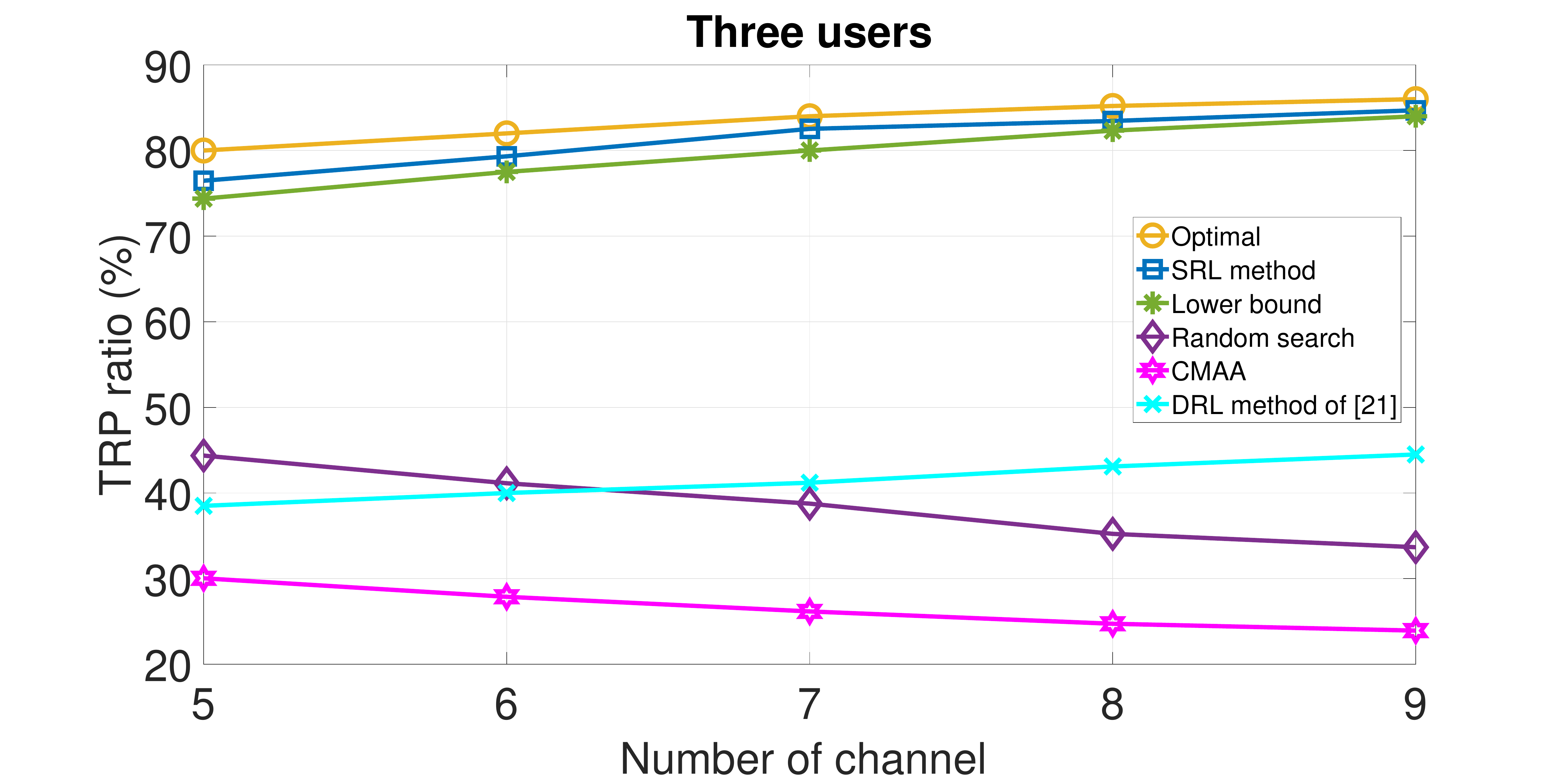} 
       \caption{  TRP for three users scenarios with channel information.}
        \label{fig8}
\end{figure}

Fig. \ref{fig8} shows the average TRP ratios of the AP  for different methods. The gaps between the TRPs result from iterative RL and the optimal ones are reduced  from $0.07\%$  to $0.03\%$ compared to the two-user scenario. This decrease is due to the availability of  the channel gains between the users and the AP   and the fact that SRL is employed. The former point helps users concentrate on exploring the optimal victim channel and power distribution which leads to more accurate solutions, while the later increases the power resolution exploration. Similar to the two users scenario, the proposed anti-jamming method outperforms the  anti-jamming methods in  \cite{Paper13}  and \cite{Paper21}. Results show that the obtained TRPs by the proposed method  are higher than the compared RL based methods with a gap more than $30\%$. The performance advantage  of our proposed RL based method  in comparison to other considered methods  shows that  the deceiving the jammer is a better policy than the others against a high-power reactive jammer. 

\begin{figure}[t]
\centering
      \includegraphics[width= 0.5\textwidth]{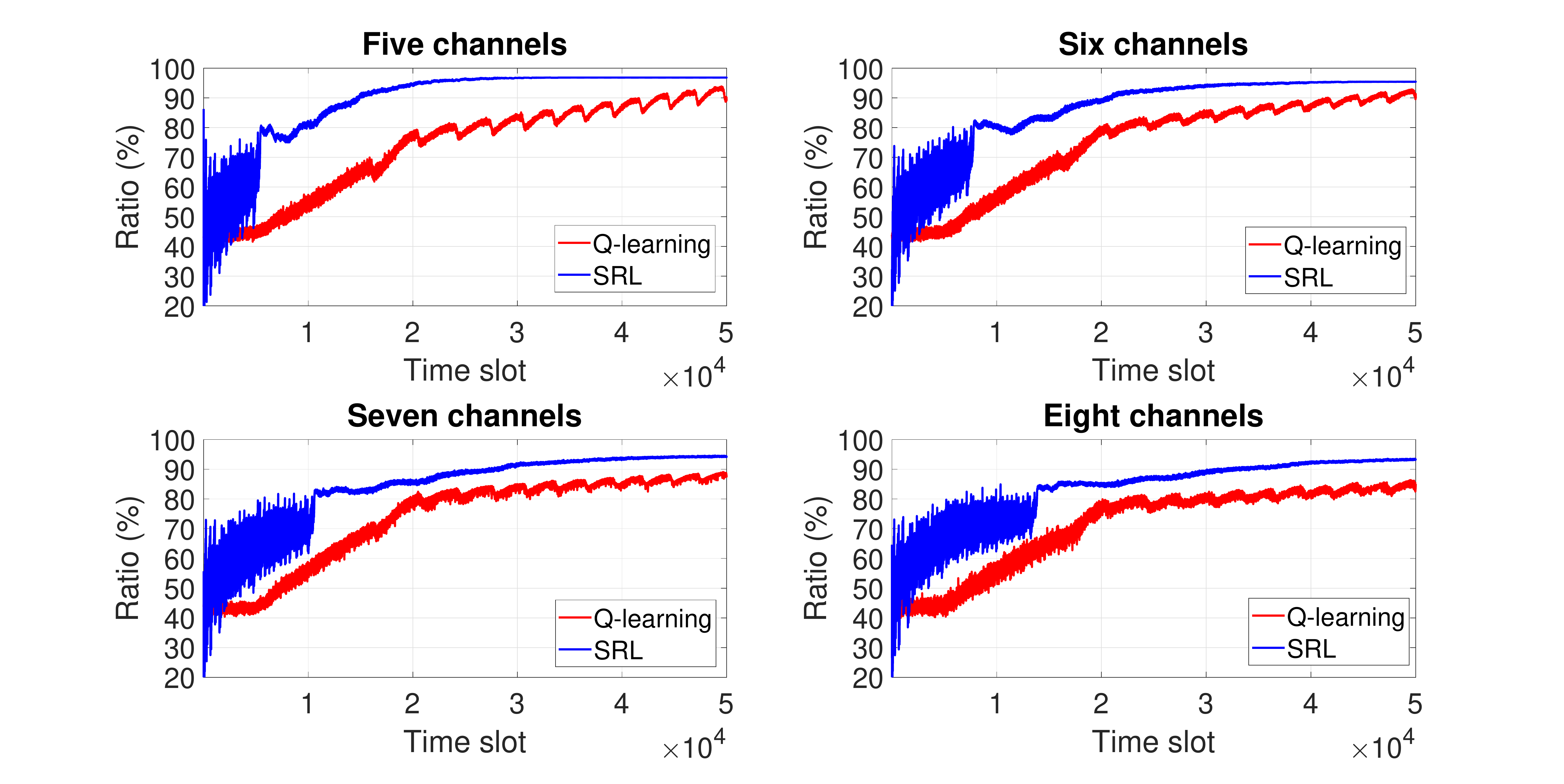}
\caption{The ratio of the AP  TRPs  to the optimal AP TRPs as a function of elapsed time slots. }
  \label{fig9}
\end{figure}

The ratios of the obtained TRP from Q-leaning and SRL methods  to the optimal AP TRP as a function of the elapsed time slots are presented in Fig. \ref{fig9}. Results are obtained  assuming three users and five to eight channels, and a power step of $0.1$. The primary power step of the SRL is set to  $2$. For all the considered cases, the proposed SRL method outperforms the Q-learning method in terms of convergence speed. Results show that SRL converges to $95\%$  of the optimal AP TRPs  for all the considered channel quantities within three iterations of RL. Moreover, the AP TRPs after obtaining the primary power allocation are over $80\%$. Hence, in the second and third iterations of RL, exploration is performed while a high AP's TRP has already been achieved. The  SLR method achieves nearly 95 percent of the optimal TRP within $20000$ time slots while Q-learning requires three times more time slots to converge to the same TRP. This is because the SRL method  reduces the number of   actions significantly. For instance, in a three-user five-channel scenario,  it is necessary to consider $101^3 \times 5$ actions for the Q-learning methods, where 101 stands for the number of power stages with a power sampling rate of 0.1, while SRL needs to explore $6^3\times 5$ actions, where 6 stands for the number of different power stages,  for the  primary resource allocation and   implementing  TD learning with $11^3$ states  two times for the final exploration.

\begin{figure}[!tbp]
\centering
 \includegraphics[width=0.5\textwidth]{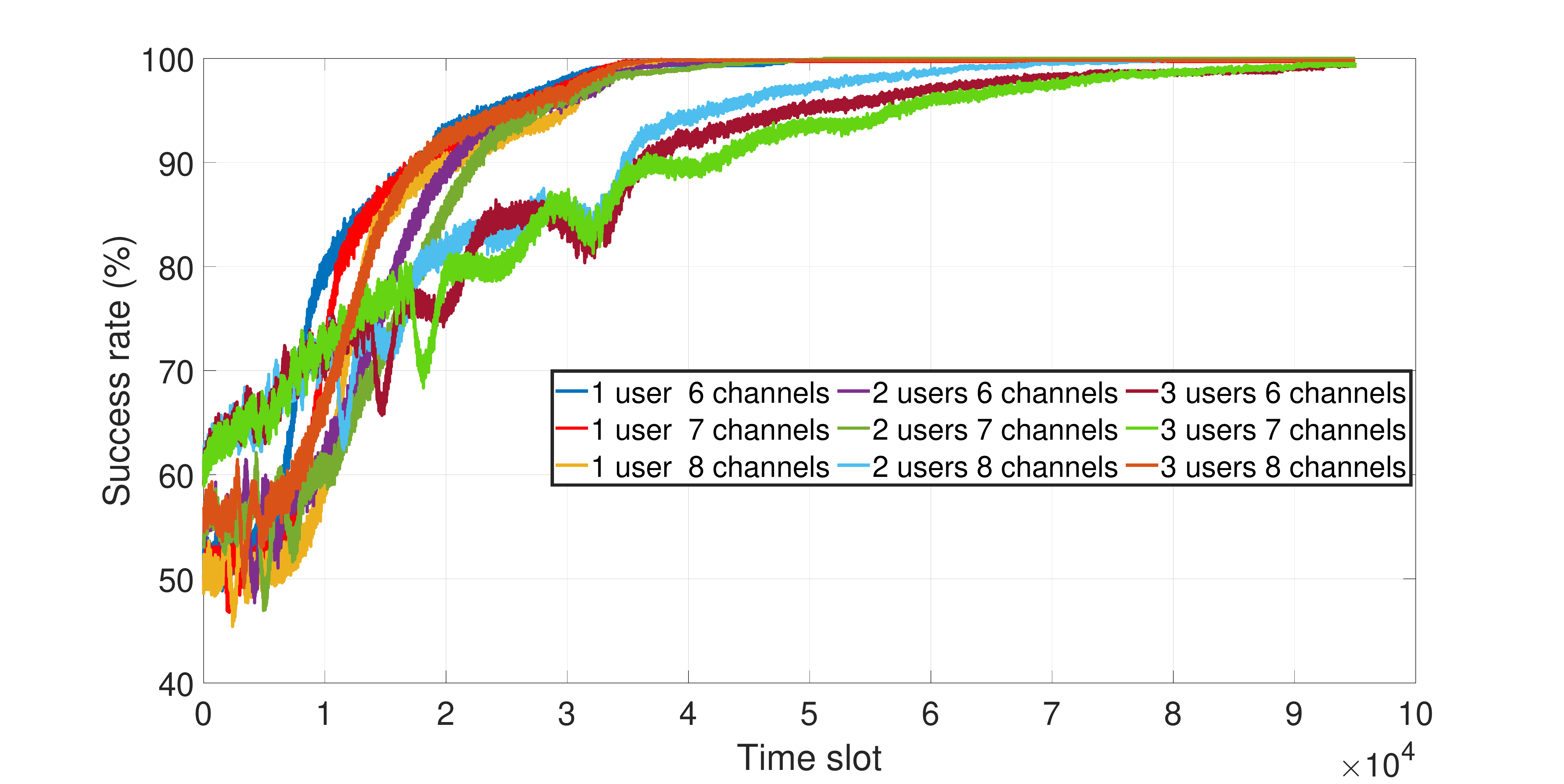}
\caption{Success rate as a function of elapsed time-slots. }
\label{fig10}
  \end{figure}
Fig. \ref{fig10}
shows the success rates for one to three users with six to eight  channels.
From this figure, we can see  that during the learning process, the jammer targets communication channels. However, as  the learning process  progresses,  the  success  rate   increases. The reason behind this is that  at the initial time-slots, the users explore  the environment to learn the optimal power and channel allocation, thus their selected power and channel are mostly  random. However, gradually, the learning process reaches  the  optimal  channel and power allocation, which provides a victim channel to deceive the jammer. After   the users  find the optimal channel and power allocation, the jammer always tends to jam the victim channel where it senses the highest signal power. The required time slots for deceiving the jammer increases  by increasing the number of users since more states and actions must be explored.

\section{Conclusion}
 
In this paper, we have proposed a  novel approach to mitigate  reactive  jamming by using a deceptive channel as a victim.  We have shown that engaging  the jammer to jam the desired victim channel  enables safe communications for legitimate users   in the other channels. To assess the proposed method, we have considered a wireless network consisting of an AP and a reactive jammer for both single-user and multi-user scenarios. Moreover, we have investigated the availability of the channel gains between the users and proposed different learning strategies  to determine the optimum resource allocation. To validate our empirical results, we have solved the power allocation problem with full knowledge of the environment and calculated a lower bound for the expectation of the total received signal. Employing the proposed method provides safe and static communication channels   for users and legitimate nodes to communicate safely with a TRP  almost equivalent  that of  the optimal solution. Moreover, the proposed SRL converges about three time faster than the RL method.

\begin{appendix}
\subsection{Proof of Proposition \ref{pro1}}
 
By using (\ref{eq15}), the matrix $\boldsymbol{M}
 $ can be represented as a sum of  matrices  $\boldsymbol{M_1}$ = $\boldsymbol{H} \cdot \boldsymbol{H} $ and $\boldsymbol{M_2}$ = $\boldsymbol{I} \cdot (\boldsymbol{h}'_j  (\boldsymbol{h}'_j)^\intercal$). Matrix $\boldsymbol{M_2}$ is positive definite  since it is a diagonal  matrix with positive elements. $\boldsymbol{M_1}$ is positive semi-definite due to the fact that it has one positive eigenvalue equal to $h_{j1}^{2}+h_{j2}^{2}+...+h_{jN}^{2}$ and $N-1$ zeros  eigenvalues. Thus, matrix $\boldsymbol{M}$ is positive definite because the sum of a positive definite and a positive semi-definite matrices is  a positive definite matrix. 

\subsection{Proof of Proposition \ref{pros2}}
 
Inequality (\ref{eq18}), for user $k$, can be reformulated as
\begin{equation}
{P_k}\leq  \bar{P} (1 - 
\frac{
\frac{\sum_{i=1}^{N}\varkappa_{ji}^{2}\xi_{ji}^{2}}{\varkappa_{jk}^{2}\xi_{jk}^{''2}}}{1+\sum_{i=1}^{N}\frac{\varkappa_{ji}^{2}\xi_{ji}^{2}}{\varkappa_{ji}^{2}\xi_{ji}^{''2}}}).
\end{equation}

Given that  $\mathbb{E}({P_k})$ = $\mathbb{E}(\mathbb{E}({P_k}|\xi_{j1},...,\xi_{jN}))$ and $E(\max \sum_{i=1}^{N}\xi_{ji}^{2}) =\Gamma$, $P_k$ can be presented as :
\begin{equation}
\begin{aligned}
&\mathbb{E}(P_k) \leq  \bar{P} (1 -  \frac{1 }{\mathbb{E}(\xi_{jk}^{''2})+\Gamma}+...+\frac{1 }{\mathbb{E}(\xi_{jk}^{''2})+\Gamma}).
\end{aligned}
\end{equation}
$\mathbb{E}(\xi_{jk}^{''2}) $ over available channels  can be obtained by $\mathbb{E}(\min(\frac{h_{jki}^{'2}}{\varkappa^2_{jk}},i\in(1,...,L))=\frac{h_{jk}^{''2}}{{\varkappa^2_{jk}}})$ as follows. If  $h'_{jil}$ is defined as the channel gain between the user $i$ and  jammer through sub-channel $l$, $F(\xi_{jk}^{''2}\leq z|(\varkappa_{j1},...,\varkappa_{jN}))$ over $N$  users and $L-1$ available channels follows
\\
\begin{equation}
\begin{aligned}
&F(\xi_{jk}^{''2}\leq z|(\varkappa_{j1},...,\varkappa_{jN}))= F(\frac{h''^2_{jk}}{\varkappa^2_{jk}}\leq z|(\varkappa_{j1},...,\varkappa_{jN}))\\&\frac{1}{N}\bigg(F(\min(\frac{h_{jkl}^{'2}}{\varkappa^2_{jk}},\forall l \in (1,...,L-1), \forall\textcolor{black}{ i \in\mathcal{X}})\leq z))+...\\&+ F(\min(\frac{h_{jkl}^{'2}}{\varkappa^2_{jk}}, \forall l \in( \textnormal{remained $L-N$ channels})\leq z))\bigg)\\&= \frac{1-(\prod_{k_1=1}^{N} e^{-\lambda(L-1)(\frac{\varkappa^2_{jk_1}}{\varkappa^2_{jk}})})}{N}+...+\frac{1-(e^{(-\lambda (L-N){z})})}{N}, 
\end{aligned}
\end{equation}
\vspace{-0.2 cm}
and 
\vspace{-0.2 cm}
\begin{equation}
\begin{aligned}
 \\&f(z)= \frac{1}{N}\bigg(\big(\lambda(L-1)(\prod_{k_1=1}^{N} e^{-\lambda(L-1)(\frac{\varkappa^2_{jk_1}}{\varkappa^2_{jk}})}\sum_{k_2=1}^N(\frac{\varkappa^2_{jk_2}}{\varkappa^2_{jk}})\big)\\&+...+\big(\lambda(L-N) e^{-\lambda(L-N) {z}}\big)\bigg),
\end{aligned}
\end{equation}
and as a result 
\begin{equation}
\begin{aligned}
\mathbb{E}(z)=\sum_{k_1=1}^{N-1}\frac{1}{\lambda N (N-k_1)(L-1-k_1)},
\end{aligned}
\end{equation}
which proves Proposition \ref{pros2}.
\vspace{-0.5 cm}
\subsection{Proof of Proposition \ref{pros3}}
Similar to the proof of Proposition \ref{pros2},  $\mathbb{E}(\xi_{ck}^{''2}) $  over available channels can be obtained using  $\mathbb{E}(\min(\frac{h_{cki}^{'2}}{\varkappa^2_{ck}},i\in(1,...,L))=\frac{h_{ck}^{''2}}{{\varkappa^2_{ck}}})$ .
 If  $h'_{cil}$ is defined as the channel gain between the user $i$ and  AP through sub-channel $l$, $F(\xi_{ck}^{''2}\leq z|(\varkappa_{c1},...,\varkappa_{cN}))$ over $N$ users and $L-1$ available channels follows
 \\
\begin{equation}
\begin{aligned}
&F(\xi_{ck}^{''2}\leq z|(\varkappa_{c1},...,\varkappa_{cN}))=\\&\frac{1}{N}\bigg(F(\max(\frac{h_{ckl}^{'2}}{\varkappa^2_{ck}},\forall l \in (1,...,L_1), \forall \textcolor{black}{ i \in\mathcal{X}})\leq z))+...\\&+F(\max(\frac{h_{ckl}^{'2}}{\varkappa^2_{ck}}, \forall l \in( \textnormal{remained $L_1-N+1$ channels})\leq z)\bigg) 
\\&=\frac{1}{N}\bigg(\big(\prod_{i=1}^{N}(1-e^{(-\lambda  \frac{\varkappa^2_{ck} }{\varkappa^2_{ci}}z)})^{L_1}\big)+\big(\prod_{i=1}^{N-1}(1-e^{(-\lambda  \frac{\varkappa^2_{ck} }{\varkappa^2_{ci}}z)})^{L_1-1}\big)\\&+...+\big((1-e^{-\lambda  {z}})^{(L_1-N+1 )}\big)\bigg),
\end{aligned}
\end{equation}
\\
and the PDF  of $z$ follows 
\\
\begin{equation}
\label{expm}
\begin{aligned}
&f(z)=\sum_{k_1=0}^{L_1}...\sum_{k_N=0}^{L_1} 
 {{L_1}\choose{k_1}}...{{L_1}\choose{k_N}}z x_1e^{-(\lambda x_1z)}(-1)^{(1+y_1)}\\&+\sum_{k_1=0}^{L_1-1}...\sum_{k_{N-1}=0}^{L_1-1} 
 {{L_1-1}\choose{k_1}}...{{L_1-1}\choose{k_{N-1}}}z x_2e^{-(\lambda x_2 z)}(-1)^{(1+y_2)}\\&+...
\sum_{k_1=0}^{L_1-N+1} 
 {{L_1-N+1}\choose{k_1}}zx_Ne^{-(\lambda x_N z)}(-1)^{(1+y_N)}.
\end{aligned}
\end{equation}
where $x_1=k_1\frac{\varkappa^2_{ck}}{\varkappa^2_{c1}}+...+k_N\frac{\varkappa^2_{ck}}{\varkappa^2_{cN}}$,  $y_1=k_1+...+k_N$,   $x_2=k_1\frac{\varkappa^2_{ck}}{\varkappa^2_{c1}}+...+k_{N-1}\frac{\varkappa^2_{ck}}{\varkappa^2_{c{N-1}}}$,  $y_2=k_1+...+k_{N-1}$,..., $x_N=k_1$, and $y_N =k_1$.
 (\ref{expm}) is sum of exponential functions having different means. Thus, $\mathbb{E}(z)$ leads to (\ref{eq24}), which proves  Proposition \ref{pros3}.
 
\subsection{ Proof of Proposition \ref{pro4}}
After determining the victim channel, the reward function  changes to
\begin{equation}
\label{eq41}
 R(\boldsymbol{d })= w_1\sum_{i=1}^{N} (\bar{P} -d^2_i )\frac{h^2_{ci}}{\bar{P}}-w_3\sum_{i=1}^{N}d^2_i, 
\end{equation}
 since the power is quantified that in Algorithm 2, we rewrite (\ref{eq41}) as a function of the allocated power from each user into the victim channel, $P_i=d^2_i $,  $i\in \mathcal{X}$. Thus, $R(\cdot)$ can be rewritten as  
\begin{equation}
 R(\boldsymbol{p })= w_1\sum_{i=1}^{N} (\bar{P} -P_i )\frac{h^2_{ci}}{\bar{P}}-w_3\sum_{i=1}^{N}P_i.
\end{equation}
Now,  assuming  that the power allocation corresponding  to the highest obtained rewards is $\boldsymbol{p^*}= [P^*_1, ..., P^*_N]$, and  $\boldsymbol{p^"}= [P^"_1, ..., P^"_N]$ is a power allocation that returns a lower reward than $\boldsymbol{p^*}$, i.e. $R(\boldsymbol{p^*} ) \geq R(\boldsymbol{p^"}  )$, if we show that for every   $\boldsymbol{\vartheta} $, where $\boldsymbol{\vartheta}= [{\vartheta_1, ..., \vartheta_N}]$, $R(\boldsymbol{p^*}+\boldsymbol{\vartheta} ) \geq R(\boldsymbol{p^"}+\boldsymbol{\vartheta} )$ holds, we can prove that the optimum point is in the neighborhood of   $\boldsymbol{p}^*$. To this end, if we prove that $R(\boldsymbol{p^*}+\boldsymbol{\vartheta} ) - R(\boldsymbol{p^"}+\boldsymbol{\vartheta} )\geq 0$ always holds, we obtain our desired result.
\begin{equation}
R(\boldsymbol{p^*}+\boldsymbol{\vartheta} ) =w_1 N -\sum_{i=1}^{N} \frac{(\bar{P}w_3 +w_1h^2_{ci})(P^*_i+\vartheta_i)}{\bar{P}},
\end{equation}
\\
\begin{equation}
\label{eq41}
R(\boldsymbol{p^"}+\boldsymbol{\vartheta} ) =w_1 N -\sum_{i=1}^{N} \frac{(\bar{P}w_3 +w_1h^2_{ci})(p^"_i+\vartheta_i)}{\bar{P}},
\end{equation}
\begin{equation}
R(\boldsymbol{p^*}+\boldsymbol{\vartheta} )-R(\boldsymbol{P^"}+\boldsymbol{\vartheta} )= \sum_{i=1}^{N} \frac{(\bar{P}w_3 +w_1h^2_{ci})(P^"_i-P^*_i)}{\bar{P}}.
\label{eq44}
\end{equation}
\\
Here, (\ref{eq44}) is always greater than zero due to the result of the considered assumption i.e.\\
($R(\boldsymbol{p^*} ) \geq R(\boldsymbol{p^"}  )$).
\begin{equation}
    R(\boldsymbol{p^*} ) \geq R(\boldsymbol{p^"}  )\rightarrow  \sum_{i=1}^{N} \frac{(\bar{P}w_3 +w_1h^2_{ci})(P^"_i-P^*_i)}{\bar{P}}\geq 0.
\end{equation}
Moreover, since the quantization step  of the power is $ \Omega_{\textrm{TD}}$, when $|\vartheta_i|\geq \Omega_{\textrm{TD}}$ , $\boldsymbol{p^"}+\boldsymbol{\vartheta} $ can be replaced by another power set with $|\vartheta|\leq \Omega_{\textrm{TD}}$. Hence, the fact that $R(\boldsymbol{p^*}+\boldsymbol{\vartheta} ) \geq  R(\boldsymbol{p^"}+\boldsymbol{\vartheta} )$ and $|\vartheta|\leq \Omega_{\textrm{TD}}$ hold, leads to the point that the optimal power set, which we denote $\boldsymbol{p}^{o }$, is obtained from $\boldsymbol{p^*} -\Omega_{\textrm{TD}} \leq \boldsymbol{p}^{o }\leq \boldsymbol{p^*} +\Omega_{\textrm{TD}}$. Moreover, in the next iteration  of SRL, the resolution is increased and $\boldsymbol{p^*} -\Omega_{\textrm{TD}} \leq \boldsymbol{p}\leq \boldsymbol{p^*} +\Omega_{\textrm{TD}}$ is covered  with a higher resolution.  As a result, since this process is valid for the further iterations of SRL, a sub-optimal  power allocation  can be achieved such that the error is bounded according to the final quantization step i.e. ($\boldsymbol{p^*} -\Omega_{\textrm{TD}}\leq \boldsymbol{p}^{o }\leq \boldsymbol{p^*} +\Omega_{\textrm{TD}}$).
 
\end{appendix}

\bibliographystyle{IEEEtran}
 \vspace{-0.3 cm}
\bibliography{references}
\begin{IEEEbiography} [{\includegraphics[width=1in,height=1.25in,clip,keepaspectratio]{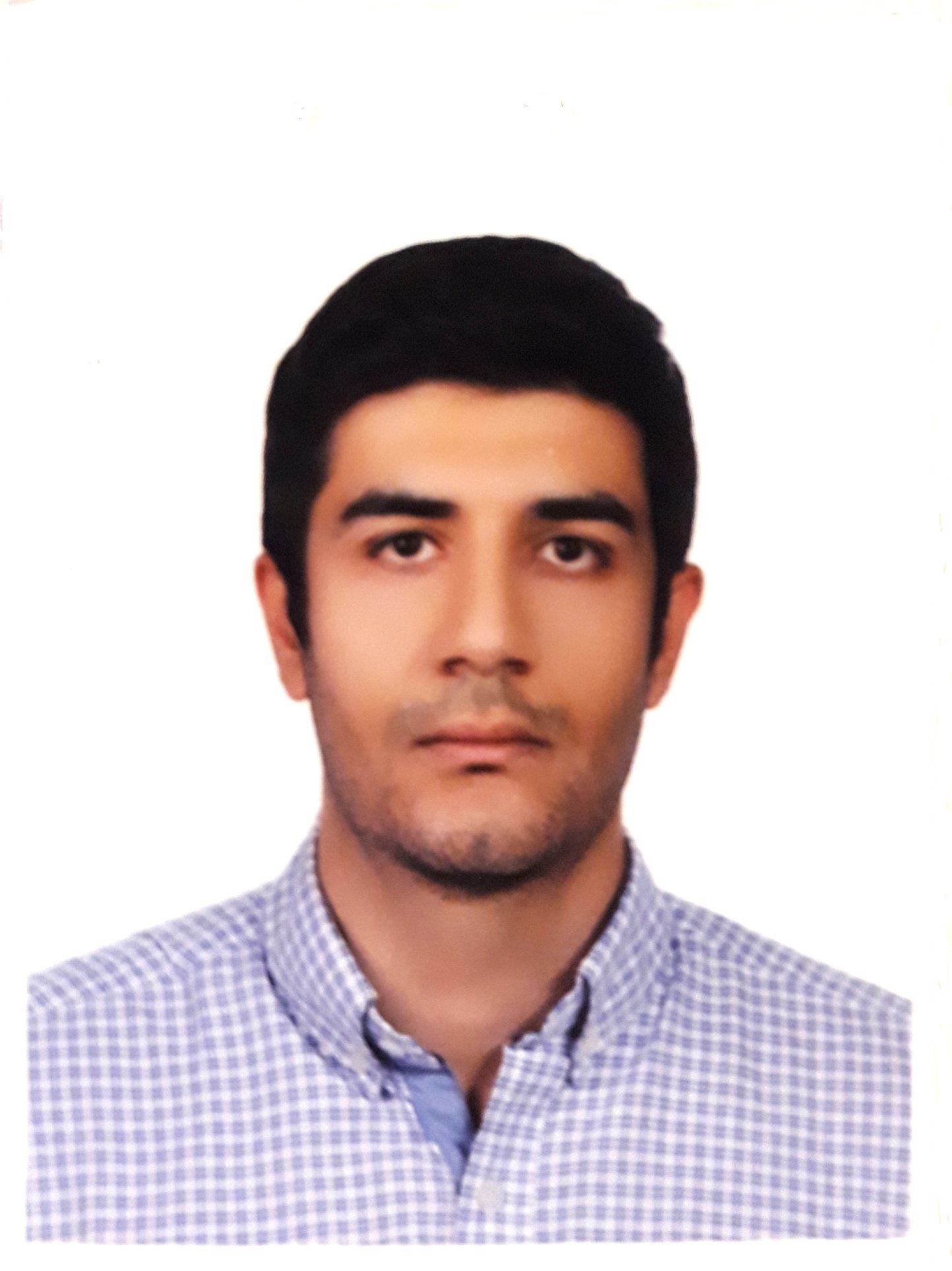}}]{Ali Pourranjbar}   received the B.S. degree in electrical engineering from International Imam Khomeini University Qazvin, Iran,
in 2011, and the M.S. degree in electrical engineering from University of Tehran, in 2015. He is currently pursuing a Ph.D. degree at the École de technologie
supérieure, Montreal, Canada. His research interests include   wireless networks, machine learning,  game  theory,  and   unmanned  aerial  vehicles.   
 \end{IEEEbiography}
 \begin{IEEEbiography} [{\includegraphics[width=1in,height=1.25in,clip,keepaspectratio]{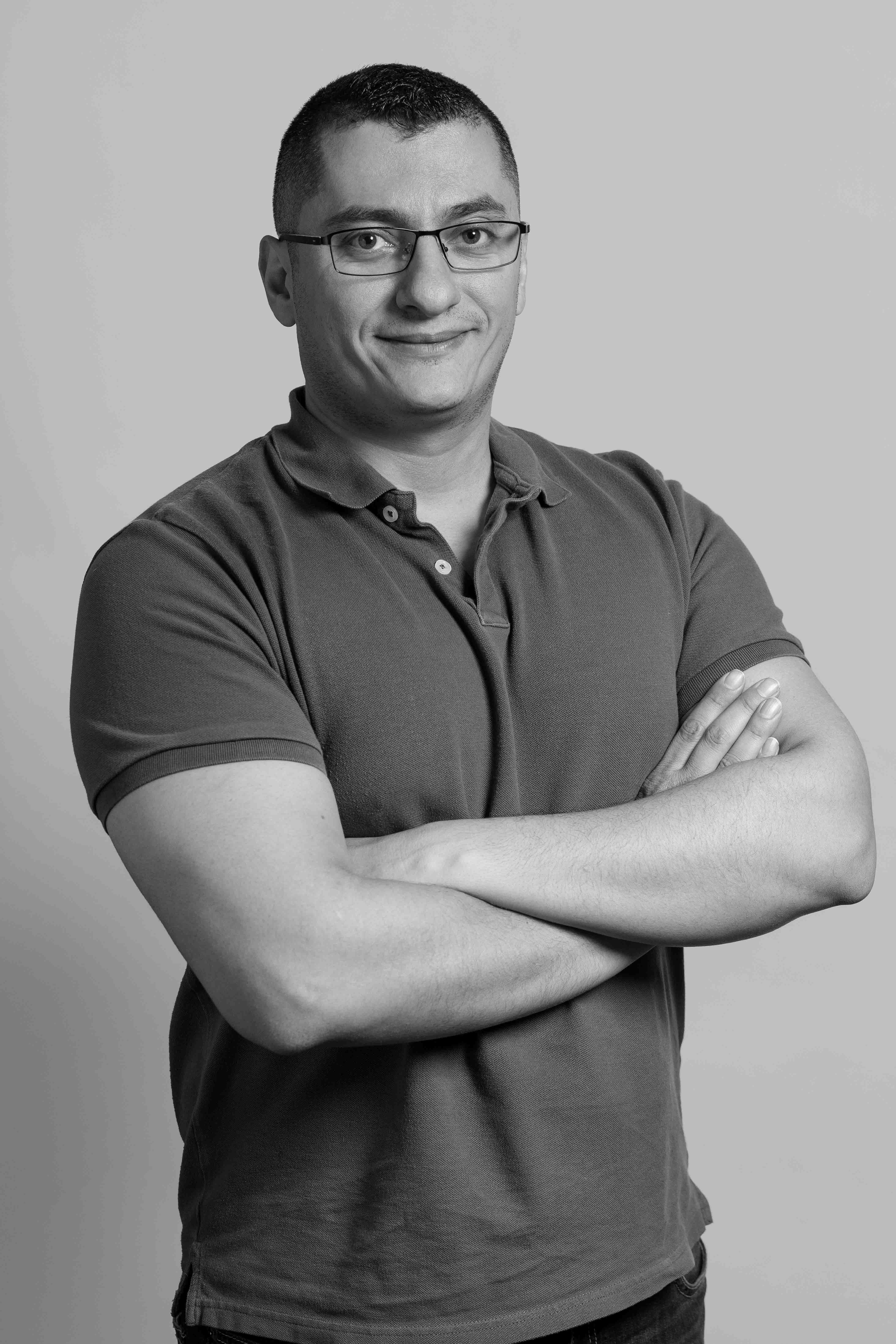}}]{Georges Kaddoum}  received the Bachelor’s degree in electrical engineering from the \'Ecole Nationale Supérieure de Techniques Avancées (ENSTA Bretagne), Brest, France, and the M.S. degree in telecommunications and signal processing(circuits, systems, and signal processing) from the Universit\'e de Bretagne Occidentale and Telecom Bretagne (ENSTB), Brest, in 2005 and the Ph.D. degree (with honors) in signal processing and telecommunications from the National Institute of Applied Sciences (INSA), University of Toulouse, Toulouse, France, in 2009. He is currently an Associate Professor and Tier 2 Canada Research Chair with the \'Ecole de Technologie Sup\'erieure (\'ETS), Universit\'e du Qu\'ebec, Montr\'eal, Canada. In 2014, he was awarded the \'ETS Research Chair in physical-layer security for wireless networks.  Since 2010, he has been a Scientific Consultant in the field of space and wireless telecommunications for several US and Canadian companies. He has published over 200+ journal and conference papers and has two pending patents. His recent research activities cover mobile communication systems, modulations, security, and space communications and navigation. Dr. Kaddoum received the Best Papers Awards at the 2014 IEEE International Conference on Wireless and Mobile Computing, Networking, Communications (WIMOB), with three coauthors, and at the 2017 IEEE International Symposium on Personal Indoor and Mobile Radio Communications (PIMRC), with four coauthors. Moreover, he received IEEE Transactions on Communications Exemplary Reviewer Award for the year 2015, 2017, 2019. In addition, he received the research excellence award of the Universit\'e du Qu\'ebec in the year 2018. In the year 2019, he received the research excellence award from the \'ETS in recognition of his outstanding research outcomes. Prof. Kaddoum is currently serving as an Associate Editor for IEEE Transactions on Information Forensics and Security, and IEEE Communications Letters. \end{IEEEbiography}
 
\begin{IEEEbiography} [{\includegraphics[width=1in,height=1.25in,clip,keepaspectratio]{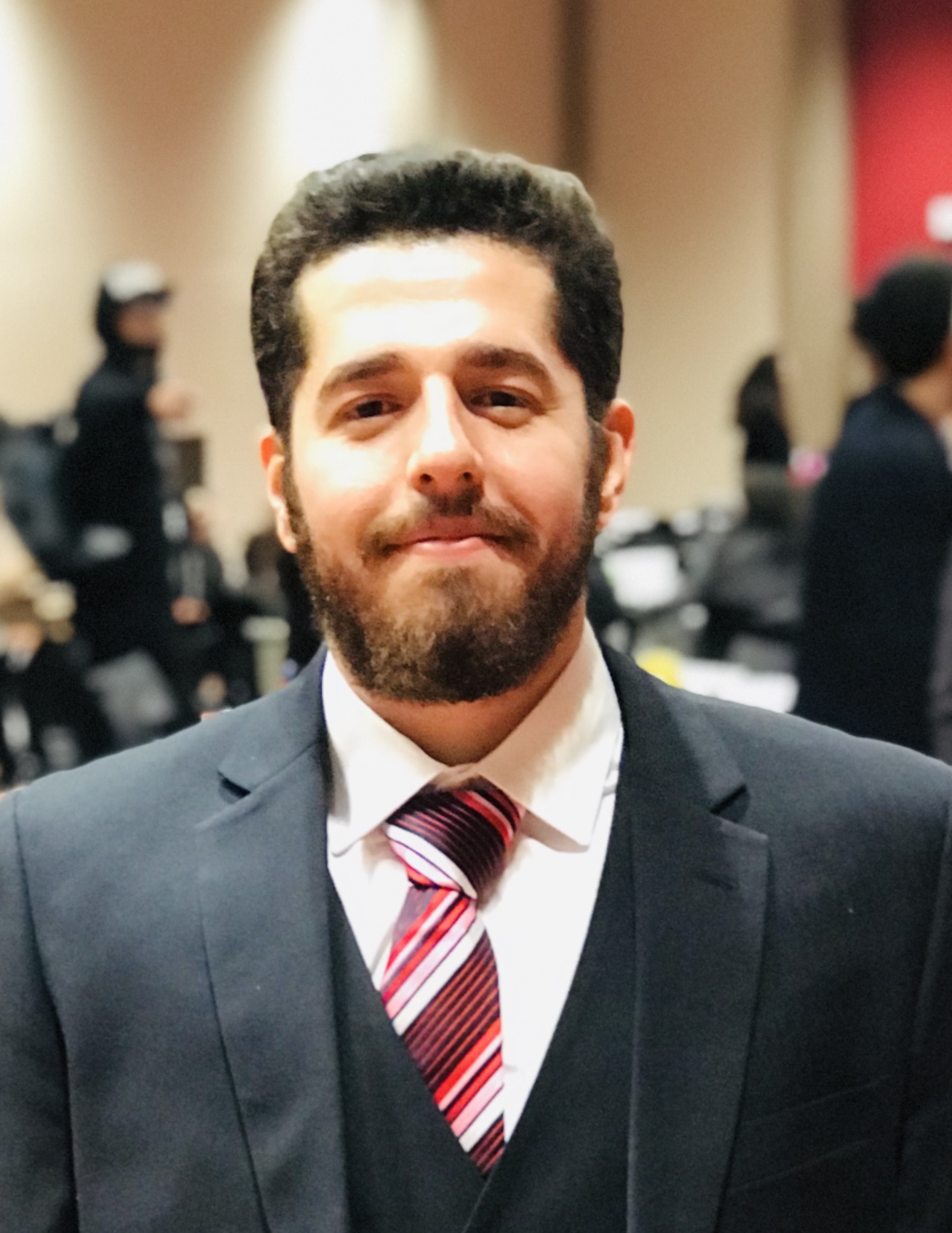}}]{Aidin Ferdowsi} (S'17) received the Ph.D. and M.S. degrees in electrical engineering from Virginia Tech and the B.S. degree in electrical engineering from the University of Tehran, Iran. He is currently a member of technical staff at Hughes Network Systems working on artificial intelligence for next-generation satellite networks.  Dr. Ferdowsi is awarded The Bill and LaRue Blackwell Graduate Research PhD Dissertation Award from Virginia Tech. He is also a fellow of Wireless@VT. His research interests include machine learning, data science, cyber-physical systems, smart cities, security, and game theory.
\end{IEEEbiography}

\begin{IEEEbiography} [{\includegraphics[width=1in,height=1.25in,clip,keepaspectratio]{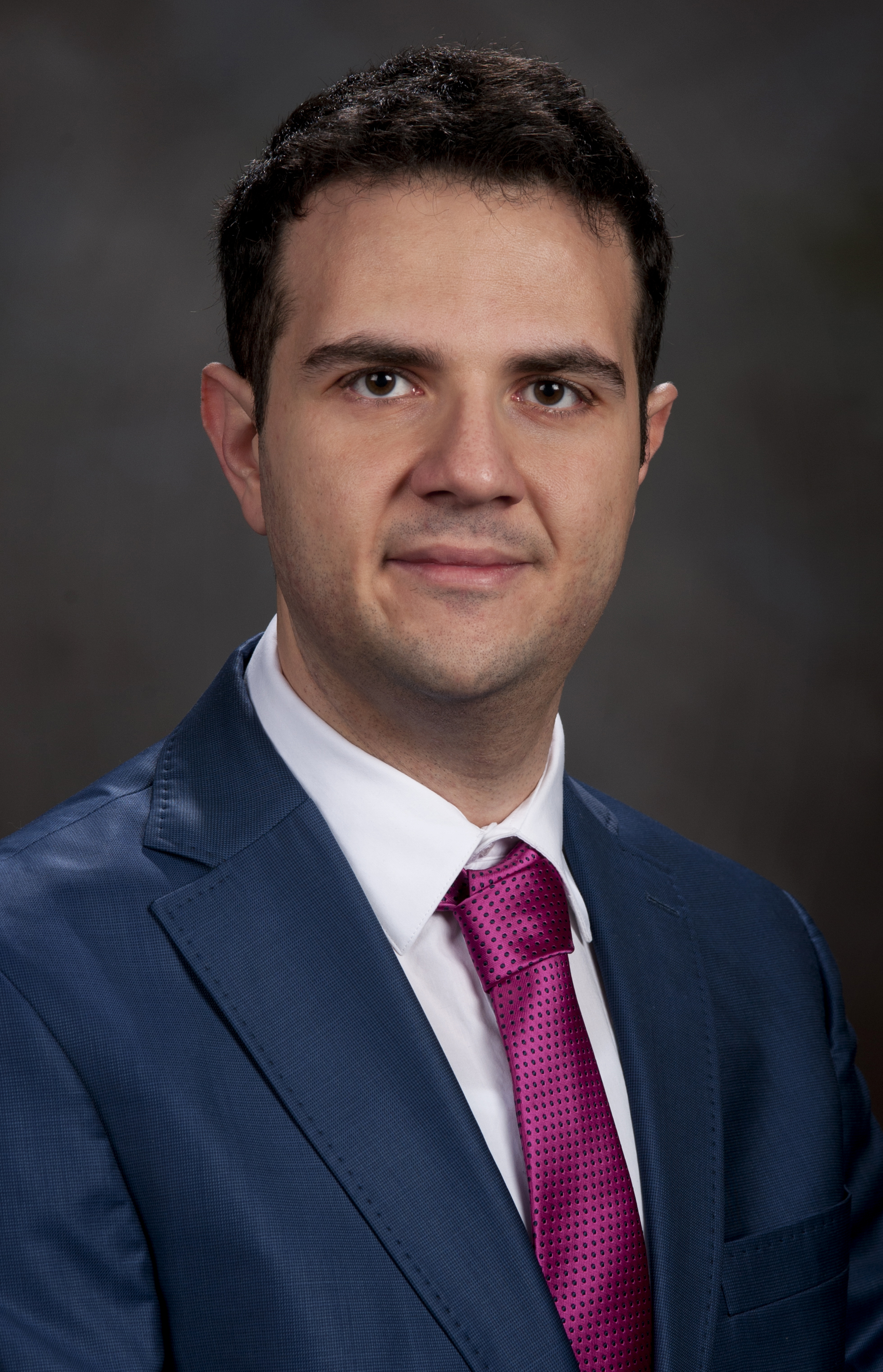}}]{Walid Saad} (S’07, M’10, SM’15, F’19) received his Ph.D degree from the University of Oslo in 2010. He is currently a Professor at the Department of Electrical and Computer Engineering at Virginia Tech, where he leads the Network sciEnce, Wireless, and Security (NEWS) laboratory. His research interests include wireless networks, machine learning, game theory, security, unmanned aerial vehicles, cyber-physical systems, and network science. Dr. Saad is a Fellow of the IEEE and an IEEE Distinguished Lecturer. He is also the recipient of the NSF CAREER award in 2013, the AFOSR summer faculty fellowship in 2014, and the Young Investigator Award from the Office of Naval Research (ONR) in 2015. He was the author/co-author of ten conference best paper awards at WiOpt in 2009, ICIMP in 2010, IEEE WCNC in 2012, IEEE PIMRC in 2015, IEEE SmartGridComm in 2015, EuCNC in 2017, IEEE GLOBECOM in 2018, IFIP NTMS in 2019, IEEE ICC in 2020, and IEEE GLOBECOM in 2020. He is the recipient of the 2015 Fred W. Ellersick Prize from the IEEE Communications Society, of the 2017 IEEE ComSoc Best Young Professional in Academia award, of the 2018 IEEE ComSoc Radio Communications Committee Early Achievement Award, and of the 2019 IEEE ComSoc Communication Theory Technical Committee. He was also a co-author of the 2019 IEEE Communications Society Young Author Best Paper. From 2015-2017, Dr. Saad was named the Stephen O. Lane Junior Faculty Fellow at Virginia Tech and, in 2017, he was named College of Engineering Faculty Fellow. He received the Dean's award for Research Excellence from Virginia Tech in 2019. He currently serves as an editor for the IEEE Transactions on Mobile Computing and the IEEE Transactions on Cognitive Communications and Networking. He is an Editor-at-Large for the IEEE Transactions on Communications. \end{IEEEbiography}
\end{document}